%% file: diversity-arxiv.tex
\documentclass{article}

\usepackage{amsthm}
\theoremstyle{plain}
\newtheorem{theorem}{Theorem}

\newtheorem{definition}[theorem]{Definition}

\usepackage[a4paper]{geometry}

\input{editors-macros.tex}

\input{dirks-macros.tex}

\author{Dirk Sudholt\\
\normalsize{}Department of Computer Science\\
\normalsize{}University of Sheffield, United Kingdom}

\title{The Benefits of Population Diversity in Evolutionary~Algorithms:\\
A Survey of Rigorous Runtime Analyses}

\begin{document}

\maketitle

\begin{abstract}{}
Population diversity is crucial in evolutionary algorithms to enable global exploration and to avoid poor performance due to premature convergence. This book chapter reviews runtime analyses that have shown benefits of population diversity, either through explicit diversity mechanisms or through naturally emerging diversity. These works show that the benefits of diversity are manifold: diversity is important for global exploration and the ability to find several global optima. Diversity enhances crossover and enables crossover to be more effective than mutation. Diversity can be crucial in dynamic optimization, when the problem landscape changes over time. And, finally, it facilitates search for the whole Pareto front in evolutionary multiobjective optimization.

The presented analyses rigorously quantify the performance of evolutionary algorithms in the light of population diversity, laying the foundation for a rigorous understanding of how search dynamics are affected by the presence or absence of population diversity and the introduction of diversity mechanisms.
\end{abstract}

\section{Introduction}
\label{ds:sec:introduction}

Evolutionary algorithms (EAs) are popular general-purpose metaheuristics inspired by the natural evolution of species. By using operators like mutation, recombination, and selection, a multi-set of solutions---the \emph{population}---is evolved over time. The hope is that this artificial evolution will explore vast regions of the search space and yet use the principle of ``survival of the fittest'' to generate good solutions for the problem at hand. Countless applications as well as theoretical results have demonstrated that these algorithms are effective on many hard optimization problems.

A key distinguishing feature from other approaches such as local search or simulated annealing is the use of a \emph{population} of candidate solutions. The use of a population allows evolutionary algorithms to explore different areas of the search space, facilitating global exploration. It also enables the use of recombination, where the hope is to combine good features of two solutions.

A common problem in evolutionary algorithms is called \emph{premature convergence}: the population collapsing to copies of the same genotype, or more generally, a set of very similar genotypes, before the search space has been explored properly. In this case there is no benefit of having a population; in the worst case, the evolutionary algorithm may behave like a local search algorithm, but with an additional overhead from maintaining many similar solutions.

What we want instead is a \emph{diverse} population that contains dissimilar individuals to promote exploration. The benefits of diversity are manifold:
\begin{description}
\item[\bf global exploration:] a diverse population is generally well suited for global exploration, as it can explore different regions of the search space, reducing the risk of the whole population converging to local optima of low fitness.
\item[\bf facilitating crossover:] often a diverse population is required for crossover to work effectively. Crossing over two very similar solutions will result in an offspring that is similar to both parents, and this effect can also be achieved by mutation. Many problems where crossover is essential do require a diverse population.
\item[\bf decision making:] a diverse population provides a diverse set of solutions for a decision maker to choose from. This is particularly important in multi-objective optimization as there are often trade-offs between different objectives, and the goal is to provide a varied set of solutions to a decision maker.
\item[\bf robustness:] a diverse population reduces the risk of getting stuck in a local optimum of bad quality. It is also robust with regards to uncertainty, like noisy fitness evaluations or changes to the fitness function in case the problem changes dynamically. A diverse population may be able to track moving optima efficiently, or to maintain individuals on different peaks, such that when the global optimum changes from one peak to another, it is easy to rediscover a global optimum.
\end{description}

In the long history of evolutionary computation, many solutions have been proposed to maintain or promote diversity. This includes controlling diversity through balancing exploration and exploitation via careful parameter tuning and designing selection mechanisms carefully, to explicit diversity-preserving mechanisms that can be embedded in an evolutionary algorithm. The latter include techniques such as eliminating duplicates, subpopulations with migration as in island models, and niching techniques that try to establish niches of similar search points, and preventing niches from going extinct. Niching techniques include fitness sharing (where similar individuals are forced to ``share'' their fitness, i.\,e.\ their real fitness is reduced during selection), clearing (where similar individuals can be ``cleared'' away by setting their fitness to a minimum value), and deterministic crowding (where offspring compete directly against their parents), just to name a few.

For more extensive surveys on diversity-preserving mechanisms we refer the reader to recent surveys by Shir~\cite{Shir2012}, \v{C}repin\v{s}ek, Liu, and Mernik~\cite{Crepinsek2013}, and Squillero and Tonda~\cite{Squillero2016}.
Details for the diversity mechanisms surveyed here will be presented in the respective sections.

Many of these techniques work on either the genotypic level, i.\,e.\ trying to create a diverse set of bit strings, or on a phenotypic level, trying to obtain different phenotypes, taking into consideration some form of mapping from genotypes to phenotypes. For instance, for functions of unitation (functions that depend on only on the number of 1-bits in the bit string), the genotype is a bit string, but the phenotype is given by the number of 1-bits. Diversity mechanisms can focus on genotypic diversity or phenotypic diversity.

Given the plethora of mechanisms to be applied, it is often not clear what the best strategy is. Which diversity mechanisms work well for a given problem, which don't, and, most importantly, \emph{why}? In particular, the effect such mechanisms have on search dynamics and performance are often not well understood.

This book chapter reviews rigorous theoretical runtime analyses of evolutionary algorithms where diversity plays a key role, in order to address these questions and to develop a better understanding of the search dynamics in the presence or absence of diversity.

The goal of runtime analysis is to estimate the random or expected time until an evolutionary algorithm has met a particular goal, by rigorous mathematical studies. Goals can include finding a global optimum, finding a diverse set of optima, or, specifically in the context of multi-objective optimization, finding the Pareto front of Pareto optimal solutions.
The results help to get insight into the search behavior of evolutionary algorithms in the presence or absence of diversity, and how parameters and explicit diversity mechanisms affect performance. They in particular highlight which diversity mechanisms are effective for particular problems, and which are ineffective. More importantly, they explain \emph{why} diversity mechanisms are effective or ineffective, and how to design the most effective evolutionary algorithms for the problems considered.

The presentation of these results is intended to combine formal theorems with informal explanations in order to make it accessible to a broad audience, while maintaining mathematical rigor. Instead of presenting formal proofs, we focus on key ideas and insights that can be drawn from these analyses. The reader is referred to the original papers for rigorous proofs and further details. In many cases we only present selected results from the surveyed papers, or results that are simplified towards special cases for reasons of simplicity, and the original works contain further and/or more general results.

%
%
%
%
%
%

The outline of this chapter is as follows. After some preliminaries in Section~\ref{ds:sec:preliminaries}, we first review the use of diversity-preserving mechanisms for enhancing global exploration in mutation-based evolutionary algorithms in Section~\ref{ds:sec:global-exploration}. Section~\ref{ds:sec:benefits-for-crossover} then reviews the benefits of diversity for the use of crossover in genetic algorithms. Section~\ref{ds:sec:dynamic} reviews the benefits of diversity mechanisms in dynamic optimization, where only few results are available to date. Section~\ref{ds:sec:diversity-based-parent-selection} presents a recent, novel approach: using diversity metrics to design parent selection mechanisms that speed up evolutionary multiobjective optimization by picking parents that are most effective for spreading the population on the Pareto front.
We finish with conclusions in Section~\ref{ds:sec:conclusions}.

This chapter is not meant to be comprehensive; in fact, there are several further runtime analyses not surveyed in this chapter. This includes island models with crossover~\cite{Neumann2011,Watson2007}, achieving diversity through heterogenous island models and how this helps for \textsc{SetCover}~\cite{Mambrini2012} or achieving diversity through a tailored population model as in the case of all-pairs shortest paths~\cite{Doerr2012}. Another popular diversity mechanism is ageing: restricting the lifespan of individuals to promote diversity. There are several runtime analyses of ageing mechanisms~\cite{HorobaJansenZarges09,JansenZarges2011a,JansenZarges2011c,Oliveto2014}, however these are being reviewed in Zarges' chapter~\cite{Zarges2017}.
This chapter focusses on single-objective optimization, though diversity is a very important topic in multi-objective optimization (see, e.\,g.\, the work by Horoba and Neumann~\cite{Horoba2010b}). We only present one recent study for multiobjective optimization in Section~\ref{ds:sec:diversity-based-parent-selection} and refer to Brockhoff's survey~\cite{Brockhoff2011} for a review of further theoretical results.

\section{Preliminaries}
\label{ds:sec:preliminaries}

The rigorous runtime analysis of randomized search heuristics is a challenging task, as these heuristics have not been designed to support an analysis. For that reason, we mostly consider bare-bones algorithms to facilitate a theoretical analysis. Furthermore, enhancing a bare-bones evolutionary algorithm with a diversity mechanism allows us to compare different diversity mechanisms in a clear-cut way, keeping the baseline algorithm as simple as possible.

The algorithms presented here do not use a specific stopping criterion, as we are interested in the random time for achieving a set goal. This time is generally referred to as run time, or running time. The most common goal is finding a global optimum, and the first hitting time of a global optimum is called optimization time. In other cases we aim to find all global optima, or the whole Pareto front in a multi-objective setting.

Gao and Neumann~\cite{Gao2014} provided an alternative approach using rigorous runtime analysis: they considered the task of maximizing diversity amongst all search points with a given minimum quality. While being a very interesting study that has inspired subsequent runtime analyses~\cite{Dang2016}, it is considered out of the scope of this survey.


%
%
%

Unless mentioned otherwise, we assume a binary search space where $n$ denotes the problem size, that is, the length of the bit string. We use $0^n$ and $1^n$ to indicate the all-zeros and the all-ones string, and, more generally, blocks of bits of some value. Further, $|x|_1$ denotes the number of ones in a bit string~$x$, and $|x|_0$ denotes the number of zeros.

In the following, we say that an event occurs \emph{with high probability} if its probability is at least $1-n^{-\Omega(1)}$, and an event occurs \emph{with overwhelming probability} if its probability is at least $1-2^{-\Omega(n^{\varepsilon})}$ for some $\varepsilon > 0$.

\section{How Diversity Benefits Global Exploration}
\label{ds:sec:global-exploration}

We first review runtime analyses where explicit diversity mechanisms were used to improve the ability of evolutionary algorithms to explore the search space, and to find global optima. The first such study was presented by Friedrich, Hebbinghaus, and Neumann~\cite{Friedrich2007}, who compared a genotypic and a phenotypic diversity mechanism on an artificially constructed problem. Here we review results from subsequent works~\cite{CovantesOsuna2017,Friedrich2009,Oliveto2014a} in Section~\ref{ds:sec:twomax} that focused on the bimodal problem \twomax instead. \twomax has a simple structure, but it is very challenging to evolve a population that contains both optima. The diversity mechanisms considered for \twomax include: avoiding genotype duplicates, avoiding fitness duplicates, deterministic crowding, fitness sharing (in two variants) and clearing.

In Section~\ref{ds:sec:island-models} we review theoretical analyses for island models~\cite{Lassig2013,Lassig2014}: subpopulations that communicate via migration, which can be run effectively on parallel hardware.

\subsection{Diversity Mechanisms on TwoMax: A Simple Bimodal Function}
\label{ds:sec:twomax}

\begin{figure}[tb]
	\centering
		\begin{tikzpicture}[domain=0:30,xscale=0.25,yscale=0.25,scale=0.6, every shadow/.style={shadow
        xshift=0.0mm, shadow yshift=0.4mm}]
          \tikzstyle{helpline}=[black,thick];
          \tikzstyle{function}=[blue,very thick];
          \draw[gray!20,line width=0.2pt,xstep=1,ystep=1] (0,0) grid (30.5,15.5);
          \draw[function] (0,15) -- (15,0) -- (30,15);
          \draw[helpline, thin, -triangle 45] (0,0) -- (0,17);
          \draw[helpline, thin, -triangle 45] (0,0) -- (32,0) node[right] {\scriptsize \#{}ones};
          \draw[helpline] (0,0) -- (0,16);
          \draw[helpline] (0,0) -- (31,0);
          \draw[helpline] (0,0) -- (-1,0) node[left] {\scriptsize $n/2$};
          \draw[helpline] (0,0) -- (0,-1) node[below] {\scriptsize $0$};
          \draw[helpline] (0,15) -- (-1,15) node[left] {\scriptsize $n$};
          \draw[helpline] (15,0) -- (15,-1) node[below] {\scriptsize $n/2$};
          \draw[helpline] (30,0) -- (30,-1) node[below] {\scriptsize $n$};
		\end{tikzpicture}
	\caption{Sketch of the function \twomax with $n=30$.}
	\label{ds:fig:twomax}
\end{figure}
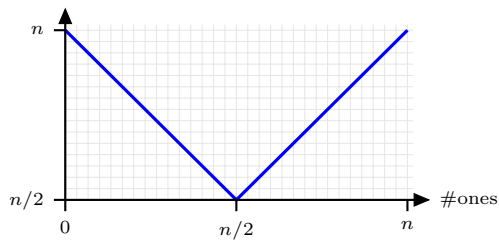

The function \twomax (see Figure~\ref{ds:fig:twomax}) is a function of unitation, that is, the fitness value only depends on the number of 1-bits:
\[
\twomax(x):=\max\left\{\sum_{i=1}^{n}x_i,n-\sum_{i=1}^{n}x_i\right\}.
\]

The fitness landscape consists of two hills with symmetric slopes: one for maximizing the number of ones, the other one for maximizing the number of zeros. These sets are also refereed to as branches. In \cite{Friedrich2009} an additional fitness value for $1^n$ was added to distinguish between a local
optimum $0^n$ and a unique global optimum. Here we use the original function with two global optima as also used in
\cite{Oliveto2014a} and \cite{CovantesOsuna2017}, and measure the time needed in order to find both optima. The presentation of results in~\cite{Friedrich2009} has been adapted to reflect this change.

\twomax is an ideal benchmark function for
studying diversity mechanisms as it is simply structured, hence facilitating a theoretical analysis, and it is hard for EAs to
find both optima as they have the maximum possible Hamming distance.
The \Twomax function does appear in well-known combinatorial
optimization problems.  For example, the \textsc{VertexCover} bipartite graph
analyzed in Oliveto, He, and Yao~\cite{Oliveto2008tr} consists of two branches, one leading
to a local optimum and the other to the minimum cover.
Another function with a similar structure is the \textsc{Mincut} instance
from Sudholt~\cite{Sudholt2010}.

\subsubsection{No Diversity Mechanism}

In order to obtain a fair comparison of different diversity-preserving mechanisms, we keep one algorithm fixed as much as possible.
The basic algorithm, the \dsmuea shown in Algorithm~\ref{ds:alg:muea}, has already been investigated by Witt~\cite{Witt2006}.

\begin{algorithm2e}[thb]
  $t \gets 0$\\
  Initialize $P_0$ with $\mu$ individuals chosen uniformly at random.\\
  \While{termination criterion not met}{
    Choose $x \in P_t$ uniformly at random.\\
    Create $y$ by flipping each bit in $x$ independently with probability $1/n$.\\
    Choose $z \in P_t$ with worst fitness uniformly at random.\\
  \eIf{$f(y) \ge f(z)$\label{ds:li:eval}}{$P_{t+1} = P_{t} \setminus \{z\}\cup \{y\}$}{$P_{t+1} = P_{t}$}
	$t \gets t+1$
}
\caption{\dsmuea}
\label{ds:alg:muea}
\end{algorithm2e}

The \dsmuea uses uniform random parent selection and elitist selection for survival. As
parents are chosen uniformly at random, the selection pressure is quite low. Nevertheless,
the \dsmuea is not able to maintain individuals on both branches for a long time. We
now show that if $\mu$ is not too large, the individuals on one branch typically
get extinct before the top of the branch is reached. Thus, the \dsmuea is unlikely to find both optima and the expected time for finding both optima is very large.

\begin{theorem}[Adapted from Theorem~1 in~\cite{Friedrich2009}]
\label{ds:the:nomechanism}
The probability that the \dsmuea with no diversity-preserving mechanism and $\mu =
o(n/\!\log n)$ finds both optima of $\Twomax$ in time $n^{n-1}$ is $o(1)$.  The expected time for finding both optima is $\Omega(n^n)$.
\end{theorem}
The proof idea is to consider the first point in time where one optimum is found. Without loss of generality, we assume that this is $0^n$. From there, one of two possible events may happen: another individual with genotype~$0^n$ can enter the population or the other optimum~$1^n$ can be found.
The former event can occur if an individual with genotype~$0^n$ is selected as parent and no bit is flipped during mutation. The more copies of $0^n$ optimum are contained in the population, the larger the probability of this event becomes. On the other hand, in order to create $1^n$, a mutation has to flip all 0-bits in the parent (and no 1-bits). If the population size is small, copies of $0^n$ tend to take over the whole population before a mutation can create $1^n$. If this happens, the \dsmuea has to flip $n$ bits to create $1^n$ from $0^n$, which has probability $n^{-n}$. Even considering $n^{n-1}$ generations, the probability of this enormous jump happening is still $o(1)$, that is, converging to 0.

\subsubsection{Avoiding Genotype Duplicates}

\begin{algorithm2e}[tbh]
  \While{termination criterion not met}{
    Choose $x \in P_t$ uniformly at random.\\
    Create $y$ by flipping each bit in $x$ independently with probability $1/n$.\\
    Choose $z \in P_t$ with worst fitness uniformly at random.\\
  \eIf{$f(y) \ge f(z)$ \textbf{\upshape{}and} $y \notin P_t$}{$P_{t+1} = P_{t} \setminus \{z\}\cup \{y\}$}{$P_{t+1} = P_{t}$}
	$t \gets t+1$
}
\caption{\dsmuea avoiding genotype duplicates}
\label{ds:alg:muea-avoiding-genotypes}
\end{algorithm2e}

A simple way to
enforce diversity within the population is not to allow genotype
duplicates. We study a mechanism used in Storch~\cite{Storch2004}, where in the ``if'' statement of the \dsmuea the condition ``and $y \notin P_t$'' is added, see Algorithm~\ref{ds:alg:muea-avoiding-genotypes}. Note that here and in the following we only show the main loop as the initialization steps are the same for all \dsmuea variants.

This mechanism ensures that the population always contains $\mu$ \emph{different} genotypes (modulo possible duplicates occurring during initialization).
However, this mechanism is not powerful enough to
explore both branches of \Twomax.
\begin{theorem}[Adapted from Theorem~2 in~\cite{Friedrich2009}]
\label{ds:the:nogenotypeduplicates}
The probability that the \dsmuea with genotype diversity and $\mu =
o(n^{1/2})$ finds both optima of $\Twomax$ in time $n^{n-2}$ is at most $o(1)$.  The expected time for finding both optima is $\Omega(n^{n-1})$.
\end{theorem}
The proof idea is similar to that of Theorem~\ref{ds:the:nogenotypeduplicates}; however we cannot rely on copies of one optimum $0^n$ taking over the population, as duplicates of $0^n$ are prevented from entering the population. The algorithm can still generate individuals similar to~$0^n$, for example by choosing $0^n$ as parent and flipping a single 0-bit to 1. As mutations can flip any of $n$ bits, the algorithm can easily create a population containing $0^n$ and many search points with only a single 1-bit that are at least as fit as the current best search points on the other branch. If the population size is $\mu = o(n^{1/2})$, the population is likely to be taken over by such search points before the other optimum~$1^n$ is found. Note that our arguments rely on $0^n$ being selected as parent, and there is only one individual with genotype~$0^n$. This leads to a more restrictive condition on $\mu$ ($\mu = o(n^{1/2})$) compared to the setting of no diversity mechanism ($\mu = o(n/\!\log n)$).

We conclude that avoiding genotype duplicates does create diversity in the population in a sense of different genotypes, but this kind of diversity is too weak for finding both optima of \twomax as there we need to evolve individuals on both branches.

\subsubsection{Fitness Diversity}

\begin{algorithm2e}[h]
  \While{termination criterion not met}{
    Choose $x \in P_t$ uniformly at random.\\
    Create $y$ by flipping each bit in $x$ independently with probability $1/n$.\\
    \eIf{there exists $z \in P_t$ such that $f(y) = f(z)$}{$P_{t+1} = P_{t} \setminus \{z\}\cup \{y\}$}{
        Choose $z \in P_t$ with worst fitness uniformly at random.\\
        \eIf{$f(y) \ge f(z)$}{$P_{t+1} = P_{t} \setminus \{z\}\cup \{y\}$}{      $P_{t+1} = P_{t}$}
    }
	$t \gets t+1$
}
\caption{\dsmuea with fitness diversity}
\label{ds:alg:muea-fitness-diversity}
\end{algorithm2e}

Avoiding genotype duplicates does not help much to optimize \Twomax as
individuals from one branch are still allowed to spread on a certain
fitness level and take over the population. A more restrictive
mechanism is to avoid fitness duplicates, i.\,e., multiple individuals
with the same fitness. Such a mechanism has been defined and analyzed
by Friedrich, Hebbinghaus, and Neumann~\cite{Friedrich2007} for plateaus of constant
fitness.
In addition, this resembles the idea of fitness diversity proposed by Hutter and Legg~\cite{Hutter2006}.

The \dsmuea with fitness diversity avoids that multiple
individuals with the same fitness are stored in the population. If at
some time~$t$ a new individual $x$ is created with the same
fitness value as a pre-existing one $y \in P_t$ then $x$ replaces~$y$ (see Algorithm~\ref{ds:alg:muea-fitness-diversity}).

The following theorem proves that if the population is not too large,
then with high probability the individuals climbing one of the two
branches will be extinguished before any of them reaches the top.
\begin{theorem}[Adapted from Theorem~3 in~\cite{Friedrich2009}]
\label{ds:the:nophenotypeduplicates2}
The probability that the \dsmuea with fitness diversity and $\mu \le
n^{O(1)}$ finds both optima\footnote{Due to the fitness diversity mechanism, and since, in contrast to~\cite{Friedrich2009}, here we consider a \twomax variant with two global optima, the population can never actually contain both optima. To set a meaningful target, here we also consider cases where the union of the current population and a new offspring contains both optima.} of $\Twomax$
 in time $2^{cn}$, $c>0$ being an appropriate
constant, is at most $o(1)$.  The expected time for finding both optima is $2^{\Omega(n)}$.
\end{theorem}
The intuitive reason why this mechanism fails is that, once the population has reached one optimum, w.\,l.\,o.\,g.\ $0^n$, the population has a tendency to spread on the branch leading to said optimum, until the whole population is contained on said branch. During this time there may be a competition between the two branches: whenever one branch creates an offspring on the same branch, it may remove an individual of the same fitness in the opposite branch. This competition is biased towards the branch at $0^n$, though, as the former branch can use ``downhill'' mutations, that is, a mutation flipping only one of the many 0-bits, whereas the opposite branch may have to rely on much rarer ``uphill'' mutations, that is, flipping only one of the rare 0\nobreakdash-bits.
The proof defines a potential function that captures the progress in this competition and shows that the branch that reaches its optimum first is likely to make individuals on the other branch go extinct.

\subsubsection{Deterministic Crowding}

\begin{algorithm2e}[tbh]
  \While{termination criterion not met}{
    Choose $x \in P_t$ uniformly at random.\\
    Create $y$ by flipping each bit in $x$ independently with probability $1/n$.\\
  \eIf{$f(y) \ge f(x)$}{$P_{t+1} = P_{t} \setminus \{x\}\cup \{y\}$}{$P_{t+1} = P_{t}$}
	$t \gets t+1$
}
\caption{\dsmuea with deterministic crowding}
\label{ds:alg:muea-deterministic-crowding}
\end{algorithm2e}

The main idea behind deterministic crowding is that offspring directly compete with their parents.
In genetic algorithms with crossover, pairs of parents are being formed, recombined, and mutated, and then the offspring competes with one of its parents, replacing it if it is no worse.

We consider this mechanism in the absence of crossover, where offspring compete with their only parent. Then the population contains $\mu$ lineages that evolve independently (see Algorithm~\ref{ds:alg:muea-deterministic-crowding}).

For sufficiently large populations the algorithm can easily reach both global optima.
\begin{theorem}[Adapted from Theorem~4 in~\cite{Friedrich2009}]
\label{ds:the:multistarts}
The \dsmuea with deterministic crowding and $\mu \le n^{O(1)}$ reaches on
$\Twomax$ a population consisting of only global optima in
expected time $O(\mu n \log n)$.  In that case the population contains
both global optima with probability at least $1 - 2^{-\mu+1}$.
\end{theorem}
The probability of $1-2^{-\mu-1}$ follows from the fact that all $\mu$ lineages evolve independently, and that for each, once a global optimum is found, $0^n$ and $1^n$ are each found with probability~$1/2$. So when fixing one lineage that reaches a global optimum, the probability that the other $\mu-1$ lineages all reach the same optimum is $2^{-\mu+1}$.
The time bound is not immediate as the \dsmuea picks a lineage to evolve further uniformly at random, so different lineages may receive different numbers of mutation steps. However, it is not difficult to show that the mutation steps are fairly concentrated around their expectation, leading to an upper time bound of $O(\mu n \log n)$.

\subsubsection{Fitness Sharing}

Fitness sharing~\cite{Mahfoud1997} derates the real
fitness of an individual $x$ by an amount that represents the
similarity of $x$ to other individuals in the population. The
similarity between $x$ and $y$ is measured by a so-called
\emph{sharing function} $\mathrm{sh}(x, y) \in [0,1]$ where a large value
corresponds to large similarities and value~0 implies no similarity.
The idea is that if there are several copies of the same individual in
the population, these individuals have to share their fitness. As a
consequence, selection is likely to remove such clusters and to keep
the individuals apart. We define the \emph{shared fitness} of~$x$ in
the population $P$ and the fitness $f(P)$ of the population, respectively, as
\[
    f(x, P) \;=\; \frac{f(x)}{\sum_{y \in P} \mathrm{sh}(x, y)} \qquad \text{and} \qquad f(P) \;=\; \sum_{x \in P} f(x, P).
\]
It is common practice to use a so-called \emph{sharing distance}
$\sigma$ such that individuals only share fitness if they have
distance less than $\sigma$. Given some distance function $d$, a
common formulation for the sharing function is
\[
    \mathrm{sh}(x, y) \;=\; \max\{0, 1 - (d(x, y)/\sigma)^\alpha\}
\]
where $\alpha$ is a positive constant that regulates the shape of the
sharing function. We use the standard setting $\alpha = 1$ and,
following Mahfoud~\cite{Mahfoud1997}, we set the sharing distance
to $\sigma = n/2$ as this is the smallest value allowing
discrimination between the two branches.  As \Twomax is a function of
unitation, we allow the distance function $d$ to depend on the number
of ones: $d(x, y) := \big||x|_1-|y|_1\big|$.  Such a strategy is
known as \emph{phenotypic sharing}~\cite{Mahfoud1997}. Our precise
sharing function is then
\[
    \mathrm{sh}(x, y) \;=\; \max\left\{0,\ 1 - 2\,\dfrac{\big||x|_1-|y|_1\big|}{n}\right\}.
\]

There are different ways of performing selection according to the shared fitness, differing in the way the reference population $P$ in the shared fitness $f(x, P)$ is chosen. In the following, we will review runtime analyses for two different variants of fitness sharing.

The most common usage of fitness sharing is to consider the shared fitness according to the union of parents and offspring, see Algorithm~\ref{ds:alg:muea-fitness-sharing}.

\begin{algorithm2e}[h]
  \While{termination criterion not met}{
    Choose $x \in P_t$ uniformly at random.\\
    Create $y$ by flipping each bit in $x$ independently with probability $1/n$.\\
    Choose $z \in P_t$ with worst fitness uniformly at random.\\
    Let $P_t' := P_t \cup \{y\}$.\\
  \eIf{$f(y, P_t') \ge f(z, P_t')$\label{ds:li:eval}}{$P_{t+1} = P_{t} \setminus \{z\}\cup \{y\}$}{$P_{t+1} = P_{t}$}
	$t \gets t+1$
}
\caption{\dsmuea with fitness sharing}
\label{ds:alg:muea-fitness-sharing}
\end{algorithm2e}

Oliveto, Sudholt, and Zarges~\cite{Oliveto2014a} showed that a population size of $\mu =2$ is not sufficient to find both optima, and that the performance is even worse than for deterministic crowding with $\mu=2$.
The following theorem states that with a probability greater than $1/2$, the (2+1)~EA will end up with both individuals in the same optimum, leading to an exponential running time from there. This performance is worse than for deterministic crowding, for which the probability of finding both optima is exactly~$1/2$ (see Theorem~\ref{ds:the:multistarts}).
\begin{theorem}[Theorems 1 and 2 in~\cite{Oliveto2014a}]
The (2+1)~EA with fitness sharing
with probability $1/2 + \Omega(1)$ will reach a population with both members in the same optimum,
and then the expected time for finding both optima from there is $\Omega(n^{n/2})$.\label{ds:the-mu2-negative}

However, with probability $\Omega(1)$ the algorithm will find both optima in time $O(n \log n)$.
\end{theorem}
The reason for the failure probability of $1/2 + \Omega(1)$ is that the algorithm typically gets stuck on one branch if both initial search points are on the same branch (which happens with probability around~$1/2$) or if the search points are initialized on different branches, but one search point has a much higher fitness than the other. In that case the effect of fitness sharing is not strong enough, and the less fit individual will be replaced if the fitter one creates an offspring similar to itself.

In case the population is initialized with two search points on different branches, and similar fitness, fitness sharing ensure that, with high probability, individuals on both branches survive. The reason is that, whenever one parent creates an offspring on its branch, fitness sharing derates the fitness of both parent and offspring in such a way that the less fit one will be removed and the individual on the opposite branch survives.

For population sizes $\mu \ge 3$ fitness sharing becomes much more effective.
\begin{theorem}[Theorem~3 in~\cite{Oliveto2014a}]
For any population size $\mu\geq 3$ the ($\mu$+1)~EA with fitness sharing will find both optima of \twomax in expected time $O(\mu n \log n)$.
\end{theorem}
The analysis reveals a very interesting behavior. In case all search points are initialized on one branch, the population starts to climb up said branch. But once a sufficiently large overall fitness value has been obtained (at the latest when two individuals have found an optimum) then these high-fitness develop a sufficient large critical mass such that the
effect of fitness sharing starts becoming evident, and the population shows a very different behavior. From this point in time on, the population starts expanding towards lower fitness values and the individuals with the smallest and the largest numbers of 1-bits always survive. While the whole population may start to climb up one branch, at some point in time the individual with the lowest fitness starts to be repelled and makes its way back down, eventually reaching the other branch and climbing up to find the other optimum.

We can conclude that fitness sharing works for the \dsmuea with population sizes $\mu \ge 3$, but when considering larger offspring populations\footnote{The \dsmlea is a variant of the \dsmuea creating $\lambda$ offspring in parallel and then selecting the $\mu$ best according to $f(\cdot, P_t')$, where $P_t'$ is the union of all $\mu$~parents and $\lambda$~offspring, breaking ties towards preferring offspring.} it can have undesirable effects: if a cluster of individuals creates too many offspring, sharing decreases the shared fitness of all individuals in the cluster, and the cluster may go extinct.

In a similar vein, the population can even lose all global optima. In a (2+$\lambda$)~EA with $\lambda \ge 6$, if the population contains 2 copies of the same global optimum, and then a generation creates $\lambda-2$ clones and 2 individuals with Hamming distance~1 to the optimum, the latter two individuals will have a higher shared fitness and form the new population.

The following result shows that even with a small offspring population size of $\lambda=2$ the \dsmlea can fail.
\begin{theorem}[Theorem~4 in~\cite{Oliveto2014a}]
With probability $1-o(1)$ 
the (2+2)~EA with fitness sharing will, at some point of time, reach a population with both members in the same optimum. The expected time for finding both optima from there is $\Omega(n^{n/2})$.\label{ds:the-largeLambda}
\end{theorem}

In order to avoid these problems, early runtime analyses of fitness sharing~\cite{Fischer2005,Friedrich2009,Sudholt2005} used fitness sharing in a different sense. They set up a competition between \emph{populations} instead of individuals: the \dsmuea variant considers the union of the parent population and the offspring population, $P_t'$ and then selects the subset $P^* \subset P_t'$ of size~$\mu$ that maximizes $f(P^*)$. This makes sense as the goal is to evolve a population of high population fitness.

Friedrich \emph{et al.}~\cite{Friedrich2009} showed that the \dsmuea with a population-level implementation of fitness sharing can efficiently find both optima on $\Twomax$.
\begin{theorem}[Adapted from Theorem~5 in~\cite{Friedrich2009}]
\label{ds:the:fitnesssharing}
    The \dsmuea with fitness sharing and $\mu \ge 2$ finds both optima on $\Twomax$ in expected time $O(\mu n \log n)$.
\end{theorem}
The reason for this efficiency is as follows. Imagining all parents and the new offspring on a scale
of the number of 1-bits, the individuals with the smallest and the largest
number of ones have the largest distance to all individuals in the
population. Therefore, fitness sharing makes these outer individuals
very attractive in terms of shared fitness, hence these individuals
are taken over to the next generation. This even holds if an outer
individual has the worst fitness in the population; the best possible population that can be formed from parents and offspring will create individuals with a minimum and a maximum number of ones.

Hence the minimum number of ones in the population can never increase, and the maximum number of ones can never decrease. Both quantities can be improved whenever the outer individuals perform a hill-climbing step towards their respective optima. Performing a hill-climbing task towards both $0^n$ and $1^n$ yields the expected time of $O(\mu n \log n)$.

A drawback of this design is that to find a population $P^*$ that maximizes the population fitness, one needs to consider up to $\binom{\mu+\lambda}{\mu}$ different candidate populations of size $\mu$. In the case of $\lambda=1$ this is $\mu+1$ combinations, but for large $\mu$ and $\lambda$ this strategy is prohibitive.

\subsubsection{Clearing}

Clearing is a niching method that uses a similar principle compared to fitness sharing. While fitness sharing can be regarded as sharing resources evenly between similar individuals, clearing assigns these resources only to the best individual of each niche. Such an individual is referred to as a \emph{winner}. All other individuals have their fitness set to 0 (or, more generally, to a value lower than the lowest fitness value in the search space).

\begin{algorithm2e}[tb]
\SetKwInOut{Input}{input}\SetKwInOut{Output}{output}
\Input{A population $P$}
\Output{Fitness values after clearing $f'$ for all $x \in P$.}
  Let $f' = f$.\\
  Sort $P$ according to decreasing $f'$ values.\\
  \For{$i := 1$ \textbf{\upshape{}to} $|P_t|$}{
    \If{$f'(P[i]) > 0$}{
        $\mathrm{winners} \gets 1$.\\
        \For{$j := i + 1$ \textbf{\upshape{}to} $|P|$}{
          \If{$f'(P[j]) > 0$ \textbf{\upshape{}and} $d(P[i], P[j]) < \sigma$}{
            \eIf{$\mathrm{winners} < \kappa$}{$\mathrm{winners} \gets \mathrm{winners} + 1$}{$f'(P[j]) \gets 0$}
          }
      }
    }
  }
\caption{Clearing procedure}
\label{ds:alg:muea-clearing-procedure}
\end{algorithm2e}

Niches are established as in fitness sharing by using a clearing radius~$\sigma$ that determines up to which distance individuals will be considered to belong to the same niche. Each niche supports up to $\kappa$ winners, where $\kappa$ is a parameter called the \emph{niche capacity}.
The decision which individuals are winners, and whose fitness is cleared, is made in a greedy procedure, shown in Algorithm~\ref{ds:alg:muea-clearing-procedure}.
The individuals are first sorted in decreasing fitness. Then the clearing procedure processes individuals in this order. For each individual, if it has not been cleared, it is declared a winner. Then procedure iterates through all remaining individuals (i.\,e.\ those with lower or equal fitness) that haven't been cleared yet and that are within a clearing distance of $\sigma$, adding them to its niche until $\kappa$ winners have been found, and clearing all remaining such individuals.

Clearing is a powerful mechanism as it allows for both exploitation and exploration: it allows winners to find fitness improvements, while at the same time enabling cleared individuals to tunnel through fitness valleys. In fact, cleared individuals are agnostic to the fitness landscape as they always have the worst possible fitness. Hence cleared individuals can explore the landscape by performing random walks. As we will show, this allows the algorithm to escape from local optima with even very large basins of attraction.
The \dsmuea with clearing is shown in Algorithm~\ref{ds:alg:muea-clearing}.


\begin{algorithm2e}[tbh]
  \While{termination criterion not met}{
    Choose $x \in P_t$ uniformly at random.\\
    Create $y$ by flipping each bit in $x$ independently with probability $1/n$.\\
    Compute the fitness $f'$ after clearing of all individuals in $P_t \cup \{y\}$ according to the clearing procedure.\\
    Choose $z \in P_t$ with worst fitness after clearing uniformly at random.\\
  \eIf{$f'(y) \ge f'(z)$\label{ds:li:eval}}{$P_{t+1} = P_{t} \setminus \{z\}\cup \{y\}$}{$P_{t+1} = P_{t}$}
	$t \gets t+1$
}
\caption{\dsmuea with clearing}
\label{ds:alg:muea-clearing}
\end{algorithm2e}


Covantes Osuna and Sudholt~\cite{CovantesOsuna2017} consider the performance of the \dsmuea with clearing for two choices for the dissimilarity measures~$d$. When choosing $d$ as the Hamming distance, we refer to this as \emph{genotypic clearing}.
Choosing the phenotypic distance as the difference in the number of ones, $d(x,y):=||x|_1-|y|_1|$ this strategy is referred to as \emph{phenotypic clearing}.

With phenotypic clearing and clearing radius~$\sigma=1$, every number of ones represents their own niche. If the population size is large enough to contain all niches, the population can easily spread throughout all the niches. In the case of \twomax, this means that both optima will have been found. In fact, this argument even extends to finding an optimum for \emph{all} functions of unitation as one of the niches will contain all global optima.
The expected time for the population to spread across all niches is $O(\mu n \log n)$, which is the same time bound as for fitness sharing and deterministic crowding.
\begin{theorem}
\label{ds:the:optunitation}
Let $f$ be a function of unitation and ${\sigma=1}$, $\mu \geq (n+1)\cdot\kappa$. Then, the expected
optimization time of the \dsmuea with phenotype clearing on $f$ is $O(\mu n\log{n})$.
\end{theorem}

For genotypic clearing we have to consider larger niches, as otherwise each niche just consists of a single search point, and genotypic clearing essentially amounts to avoiding duplicates in the population (see Theorem~\ref{ds:alg:muea-avoiding-genotypes}). The most natural choice is $\sigma=n/2$ as for fitness sharing, as this is allows to distinguish the two branches. For this setting we have the following performance guarantee.
\begin{theorem}
\label{ds:the:few-niches-twomax}
The expected time for the \dsmuea with genotypic or phenotypic clearing, $\mu \ge \kappa n^2/4$,
$\mu \le n^{O(1)}$ and $\sigma=n/2$ finding both optima on \twomax is $O(\mu n \log n)$.
\end{theorem}
The idea behind the proof is to consider the situation after one of the optima has been reached, and once the population contains $\kappa$ copies of said optimum. It is easy to show that the expected time until this happens, or both optima are being found, is bounded by $O(\mu n \log n)$. 

We then consider a potential function that describes the state of the current population: the sum of Hamming distances of all individuals to the optimum. Note that phenotypic and genotypic distances to an optimum ($0^n$ or $1^n$) are the same, hence the analysis holds for both phenotypic and genotypic clearing. Imagine the situation when all individuals are close to the optimum. Then any mutation is likely to create an offspring that is further away from the optimum. Thus, mutation has a tendency to increase the potential.

Selection will then remove one of the non-winner individuals uniformly at random. There is a small bias introduced by selection towards remaining close to the winner. This is down to the fact that losers in the population do not evolve in complete isolation. The population always contains $\kappa$ copies of the winner that may create offspring and may prevent the population from venturing far away from it. In other words, there is a constant influx of search points descending from winners.

All in all, mutation and selection yield opposite biases. The bias induced by selection decreases as the fraction of winners $\kappa/\mu$ decreases. If the population size $\mu$ is large enough with respect to $\kappa$ and $n$: $\mu \ge \kappa n^2/4$, the potential shows a positive expected change until it reaches a value from which, by the pigeon-hole principle, we can conclude that at least one individual must have reached a distance at least~$n/2$ from the winners.

From there, a new niche is being created, and the other optimum can easily be found by hill climbing. The overall time is bounded by $O(\mu n \log n)$.

Note that the condition $\mu \ge \kappa n^2/4$ is a sufficient condition, and a quite steep requirement compared to fitness sharing, which works with constant population sizes~$\mu$. On the other hand, clearing works with genotypic distances whereas fitness sharing was only proved to work with phenotypic distances. A further advantage for clearing is that it also works on variants of \twomax with different slopes, whereas the analysis of fitness sharing is sensitive to the absolute fitness values.

\subsection{Diversity in Island Models}
\label{ds:sec:island-models}

The presentation in this subsection is partly taken from this author's theory-flavored survey of parallel evolutionary algorithms~\cite{Sudholt2012a}.

Island models are popular ways of parallelizing evolutionary algorithms: they consist of subpopulations that may be run on different cores, and that coordinate their searches by using migration: communicating selected search points, or copies thereof, to other islands. These solutions are then considered for inclusion on the target island in a further selection process. Island models communicate along a communication topology, a directed graph that connects the islands, and migration involves sending solutions to all neighboring islands. Often periodic migration is used: migration happens every $\tau$ iterations, where $\tau$ is a parameter called the \emph{migration interval}.

This way, islands can communicate and compete with one another. Islands that got stuck in low-fitness regions of the search space can be taken over by individuals from more successful islands. This helps to coordinate search, focus on the most promising regions of the search space, and use the available resources effectively. They also act as an implicit diversity mechanism: between migrations, islands evolve independently, and the flow of genetic information in the whole system is slowed down, compared to having one large population. This can help to increase diversity and to prevent or at least delay premature convergence. Note that the flow of information can be tuned by tuning the migration interval~$\tau$, the migration topology, and other parameters like the number of individuals to be migrated or the policies (selection schemes) for emigration and immigration.
Algorithm~\ref{ds:alg:island-model} shows a general scheme of a basic island model.

\begin{algorithm2e}[tbh]
\SetKwFor{ForInParallel}{for}{do in parallel}{endfor}
  Initialize a population made up of subpopulations or islands, $P^{(0)} = \{P_1^{(0)}, \dots, P_{\numislands}^{(0)}\}$.\\
  $t \gets 1$.\\
  \While{termination criterion not met}{
\ForInParallel{each island~$i$}{
      \If{$t \bmod \tau = 0$}{
        Send selected individuals from island $P^{(t)}_i$ to selected neighboring islands.\\
        Receive immigrants $I_i^{(t)}$ from islands for which island $P_i^{(t)}$ is a neighbor.\\
        Replace $P_i^{(t)}$ by a subpopulation resulting from a selection among $P_i^{(t)}$ and $I_i^{(t)}$.\\
      }
      Produce $P_i^{(t+1)}$ by applying reproduction operators and selection to~$P_i^{(t)}$.\\
    }
	$t \gets t+1$
  }
\caption{Scheme of an island model with migration interval~$\tau$}
\label{ds:alg:island-model}
\end{algorithm2e}

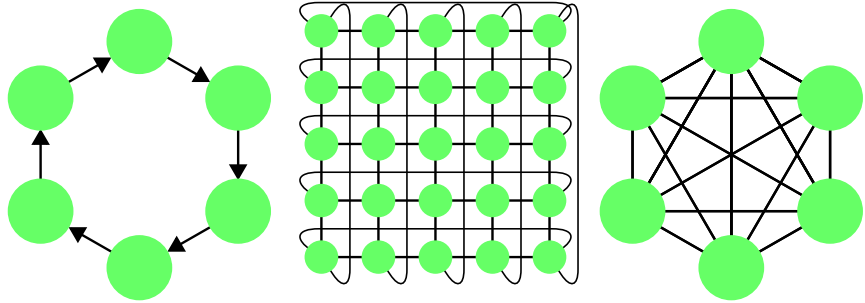
\begin{figure}[hbt]
\begin{center}
\scalebox{0.75}{
\begin{tikzpicture}[scale=2]
\tikzstyle{topology edge}=[edge,very thick,triangle 60-,shorten <=12pt]
    \foreach \u in {0,...,6} {
            \node (vertex\u) at ($(\u*360/6+90:1)$) {};
    }
            \draw[topology edge] (vertex0) edge (vertex1);
            \draw[topology edge] (vertex1) edge (vertex2);
            \draw[topology edge] (vertex2) edge (vertex3);
            \draw[topology edge] (vertex3) edge (vertex4);
            \draw[topology edge] (vertex4) edge (vertex5);
            \draw[topology edge] (vertex5) edge (vertex0);
    \foreach \u in {0,...,5} {
        \filldraw[vertex,uninformed] (vertex\u.center) circle (8pt);
    }
\end{tikzpicture}
\begin{tikzpicture}[scale=1]
\tikzstyle{topology edge}=[edge,very thick,-]
    \foreach \u in {0,...,4} {
        \draw[topology edge] (\u, 0) -- +(0, 4);
        \draw[topology edge] (0, \u) -- +(4, 0);
        \draw[topology edge, rounded corners=15pt, thick] (\u, 0) -- ++(0.5, -0.75) -- ++(0, 4+1.5) -- ++(-0.5, -0.75);
        \draw[topology edge, rounded corners=20pt, thick] (0, \u) -- ++(-0.75, 0.5) -- ++(4+1.5, 0) -- +(-0.75, -0.5);
    }

    \foreach \u in {0,...,4} {
     \foreach \v in {0,...,4} {
            \node (vertex\u) at ($(\u,\v)$) {};
            \filldraw[vertex,uninformed] (vertex\u.center) circle (8pt);
        }
    }
\end{tikzpicture}
\begin{tikzpicture}[scale=2]
\tikzstyle{topology edge}=[edge,very thick,-]
    \foreach \u in {0,...,6} {
            \node (vertex\u) at ($(\u*360/6+90:1)$) {};
    }
    \foreach \u in {0,...,6} {
        \foreach \v in {0,...,6} {
            \draw[topology edge] (vertex\u) edge (vertex\v);
        }
    }
    \foreach \u in {0,...,5} {
        \filldraw[vertex,uninformed] (vertex\u.center) circle (8pt);
    }
\end{tikzpicture}
}
\end{center}
\caption{Sketches of common topologies: a unidirectional ring, a torus, and a complete graph. Other common topologies include bidirectional rings where all edges are undirected and grid graphs where the edges wrapping around the torus are removed.}
\label{ds:fig:common-topologies}
\end{figure}

Common topologies include unidirectional rings (a ring with directed edges only in one direction), bidirectional rings, torus or grid graphs, hypercubes, scale-free graphs~\cite{DeFelice2011}, random graphs~\cite{Giacobini2005}, and complete graphs. Figure~\ref{ds:fig:common-topologies} sketches some of these topologies. An important characteristic of a topology~$T = (V, E)$ is its \emph{diameter}\index{diameter}: the maximum number of edges on any shortest path between two vertices. Formally, $\diam(T) = \max_{u, v \in V} \mathrm{dist}(u, v)$ where $\mathrm{dist}(u, v)$ is the graph distance, the number of edges on a shortest path from~$u$ to~$v$. The diameter gives a good indication of the time needed to propagate information throughout the topology. Rings and torus graphs have large diameters, while hypercubes, complete graphs, and many scale-free graphs have small diameters.

\subsubsection{A Royal Road for Island Models}

L{\"a}ssig and Sudholt~\cite{Lassig2010,Lassig2013} presented a first example where communication makes the difference between exponential and polynomial running times, in a typical run. They constructed a family of problems called $\mathrm{LOLZ}_{n,z,b,\ell}$ where a simple island model, all islands running \dsea{}s, finds the optimum in polynomial time, with high probability. This holds for a proper choice of the migration interval and any migration topology that is not too sparse.
Contrarily, both a single, large population as in the \dsmuea as well as independent islands (each running a \dsea, or even when they also run \dsmuea{}a) need exponential time, with high probability.

The basic idea of this construction is as follows. First imagine a bit string where the fitness describes the length of the longest prefix of bits with the same value. Generally, a prefix of $i$ leading ones yields the same fitness as a prefix of $i$ leading zeros, e.\,g., $111010$ and $000110$ both have fitness~3. However, the maximum possible fitness that can be attained by leading zeros is capped at some threshold value~$z$. This means that---in the long run---gathering leading ones is better than gathering leading zeros. The former leads to an optimal value, while the latter leads to a local optimum that is hard to escape from.

The effect on an EA is as follows. In the beginning the EA typically has to make a decision whether to collect leading ones (LO) or leading zeros (LZ). This not only holds for the \dsea but also for a (not too large) panmictic population as genetic drift will lead the whole population to either leading ones or leading zeros.
After a significant prefix has been gathered, this decision gradually becomes irreversible as many bits in the prefix need to be flipped at the same time to switch from leading ones to leading zeros or vice versa. So, with probability close to 1/2 the EA will end up finding an optimum by gathering leading ones, and again with probability close to 1/2 its population gets stuck in a hard local optimum.

To further increase the difficulty for EAs, this construction is repeated on several blocks of the bit string that need to be optimized one-by-one. Each block has length~$\ell$. Only if the right decision towards leading ones is made on the first block, the block can be filled with further leading ones. Once the first block contains only leading ones, the fitness depends on the prefix in the second block, and a further decision between leading ones and leading zeros needs to be made. Only if an EA makes all decisions correctly, it can find a global optimum. Table~\ref{ds:fig:LOLZ} illustrates the problem definition.

\begin{table}[tbh]
\begin{center}
\begin{tabular}{ccccc@{\hspace{1cm}}l}
$x_1$ & \color{red}{1111}\color{black}0011 & 11010100 & 11010110 & 01011110 & $\mathrm{LOLZ}(x_1) = 4$\\
$x_2$ & \color{red}{11111111} & \color{red}{11}\color{black}010100 & 11010110 & 01011110 & $\mathrm{LOLZ}(x_2) = 10$\\
$x_3$ & \color{red}{11111111} & \color{red}{11111111} & \color{red}{000}\color{black}00110 & 01011110 & $\mathrm{LOLZ}(x_3) = 19$\\
\end{tabular}
\end{center}
\caption{Examples of solutions for the function $\mathrm{LOLZ}$ with four blocks and $z=3$, along with their fitness values. All blocks have to be optimized from left to right. The sketch shows in red all bits that are counted in the fitness evaluation. Note how in $x_3$ in the third block only the first $z=3$ zeros are counted. Further 0-bits are ignored. The only way to escape from this local optimum is to flip all $z$ 0-bits in this block simultaneously.}
\label{ds:fig:LOLZ}
\end{table}

So, the problem requires an EA to make several decisions in succession. The number of blocks, $b$, is another parameter that determines how many decisions need to be made. Panmictic populations will sooner or later make a wrong decision and get stuck in some local optimum. If $b$ is not too small, the same holds for independent runs. The results from~\cite{Lassig2013} are summarized as follows; the second statement follows from the first one and the union bound.
\begin{theorem}
\label{ds:the:PanmicticModel}
Consider the \dsmuea with $\mu \le cn/(\log n)$ for an arbitrary constant~$c > 0$ on $\textsc{LOLZ}_{n, z, b, \ell}$ with $z = \omega(\log b)$, $b\ell \le n$, and $z < \ell$.
With probability at least $1-e^{-\Omega(z)}-2^{-b}$
the \dsmuea does not find a global optimum within $n^{z/3}$ generations.

The same holds when considering $s$ independent subpopulations, each running a \dsea or \dsmuea as specified above; then the probability bound becomes $1-s e^{-\Omega(z)}-s 2^{-b}$.
\end{theorem}

However, an island model can effectively communicate the right decisions on blocks to other islands. Islands that got stuck in a local optimum can be taken over by other islands that have made the correct decision. These dynamics make up the success of the island model as it can be shown to find global optima with high probability. A requirement is, though, that the migration interval is carefully tuned so that migration only transmits the right information. If migration happens before the symmetry between leading ones and leading zeros is broken, it might be that islands with leading zeros take over islands with leading ones.
We further need the topology to be able to spread the right information quickly enough: a topology $G$ is called \emph{well-expanding} if there is a constant~$\varepsilon > 0$ such that the following holds.
For every subset $V' \subseteq V$ with $|V'| \le |V|^\varepsilon$ we have $|N(V')| \ge (2+\varepsilon) |V'|$.
L{\"a}ssig and Sudholt~\cite{Lassig2013} give the following result.
\begin{theorem}
\label{ds:the:IslandModel}
Consider an island model where each island runs a \dsea with migration on a well-expanding migration topology with $\tau = n^{5/3}$ and $\mu = n^{\Theta(1)}$ subpopulations, accepting a best search point among all immigrants and the resident individual.
Let the function $\textsc{LOLZ}_{n, z, b, \ell}$ be parameterized according to $\ell = 2\tau/n = 2n^{2/3}$, $z = \ell/4 = n^{2/3}/2$, and $b \le n^{1/6}/16$.
If the migration counter~$t$ starts at $\tau/2 = n^{5/3}/2$ then with overwhelming probability the algorithm finds a global optimum
within $O(b\ell n) = O(n^2)$ generations.
\end{theorem}
The analysis is quite technical, but the main ideas can be summarised as follows. All islands optimize \textsc{LOLZ} by fixing bits from left to right, and at approximately the same pace. The migration interval is tuned such that between two migrations all islands will be starting to optimize the same new block, excluding islands that got stuck on previous blocks. Assume for the moment that islands make decisions independently on the new block (we'll discuss this assumption below). Then in expectation, half the islands will get stuck in local optima, reducing the number of ``good'' islands still on track towards finding the global optimum. Once this number has dropped below $|V|^\varepsilon$, the properties of the topology enure that these ``good'' islands propagate their information to sufficiently many other islands (many of which will be stuck in local optima) to ensure that a critical mass of ``good'' islands always survives, until a global optimum is found eventually.

An interesting finding is how islands do in fact make independent, or near-independent decisions on new blocks. After all, during migration, genetic information about all future blocks is transmitted. Hence, after migration many islands share the same genotype on all future blocks. This is a real threat as this dependence might imply that all islands make the same decision after moving on to the next block, compromising diversity.

However, under the conditions for the migration interval there is a period of independent evolution following migration, before any island moves on to a new block. During this period of independence, the genotypes of future blocks are subjected to random mutations, independently for each island. After some time the distribution of bits on these future blocks will resemble a uniform distribution. This shows that independence can be gained by periods of independent evolution. One could say that the island model combines the advantages of two worlds: independent evolution and selection pressure through migration. The island model is only successful because it can use both migration and periods of independent evolution.

\subsubsection{Island Models for Eulerian Cycles}

We also give a simple and illustrative example from combinatorial optimization to show how island models can be beneficial through providing diversity. L{\"a}ssig and Sudholt~\cite{Lassig2014} considered island models for the Eulerian cycle problem. Given an undirected Eulerian graph, the task is to find a Eulerian cycle, i.\,e., a traversal of the graph on which each edge is traversed exactly once. This problem can be solved efficiently by tailored algorithms, but it served as an excellent test bed for studying the performance of evolutionary algorithms~\cite{Doerr2007,Doerr2007a,Doerr2007e,Neumann2008d}.

Instead of bit strings, the problem representation by Neumann~\cite{Neumann2008d} is based on permutations of the edges of the graph. Each such permutation gives rise to a \emph{walk}: starting with the first edge, a walk is the longest sequence of edges such that two subsequent edges in the permutation share a common vertex. The walk encoded by the permutation ends when the next edge does not share a vertex with the current one. A walk that contains all edges represents a Eulerian cycle. The length of the walk gives the fitness of the current solution.

\begin{figure}[tb]
\centerline{
\scalebox{0.75}{
\begin{tikzpicture}[yscale=1]
\newcounter{next}
\tikzstyle{vertex}=[circle,draw,fill=black,scale=0.5]
\tikzstyle{walk}=[very thick, blue]
\foreach \z/\lab in {-1/l, +1/r} {
    \foreach \x in {0, ..., 8} {
        \node[vertex] (v\lab\x) at ($(\z, 0) +(\x*360/8:1)$) {};
    }
    \foreach \x in {0, ..., 7} {
        \pgfmathsetcounter{next}{\x+1}
        \draw (v\lab\x) edge (v\lab\thenext);
    }
}
\node[left] at (0, 0) {$v^*$};
\draw[walk] (vl3) -- (vl4) -- (vl5) -- (vl6) -- (vl7) -- (vl0);
\draw[line width=3pt,-triangle 60,shorten <=80] (0, 0) -- +(-3.5,-2.5);
\draw[line width=3pt,-triangle 60,shorten <=50] (0, 0) -- +(0,-2.75);
\draw[line width=3pt,-triangle 60,shorten <=80] (0, 0) -- +(+3.5,-2.5);
\end{tikzpicture}}}
\medskip
\centerline{
\scalebox{0.75}{
\begin{tikzpicture}[yscale=1]
\tikzstyle{vertex}=[circle,draw,fill=black,scale=0.5]
\tikzstyle{walk}=[very thick, blue]
\foreach \z/\lab in {-1/l, +1/r} {
    \foreach \x in {0, ..., 8} {
        \node[vertex] (v\lab\x) at ($(\z, 0) +(\x*360/8:1)$) {};
    }
    \foreach \x in {0, ..., 7} {
        \pgfmathsetcounter{next}{\x+1}
        \draw (v\lab\x) edge (v\lab\thenext);
    }
}
\node[left] at (0, 0) {$v^*$};
\draw[walk] (vl3) -- (vl4) -- (vl5) -- (vl6) -- (vl7) -- (vl0) -- (vr4) -- (vr3);
\end{tikzpicture}
\hspace*{1cm}
\begin{tikzpicture}[yscale=1]
\tikzstyle{vertex}=[circle,draw,fill=black,scale=0.5]
\tikzstyle{walk}=[very thick, blue]
\foreach \z/\lab in {-1/l, +1/r} {
    \foreach \x in {0, ..., 8} {
        \node[vertex] (v\lab\x) at ($(\z, 0) +(\x*360/8:1)$) {};
    }
    \foreach \x in {0, ..., 7} {
        \pgfmathsetcounter{next}{\x+1}
        \draw (v\lab\x) edge (v\lab\thenext);
    }
}
\node[left] at (0, 0) {$v^*$};
\draw[walk] (vl3) -- (vl4) -- (vl5) -- (vl6) -- (vl7) -- (vl0) -- (vr4) -- (vr5);
\end{tikzpicture}
\hspace*{1cm}
\begin{tikzpicture}[yscale=1]
\tikzstyle{vertex}=[circle,draw,fill=black,scale=0.5]
\tikzstyle{walk}=[very thick, blue]
\foreach \z/\lab in {-1/l, +1/r} {
    \foreach \x in {0, ..., 8} {
        \node[vertex] (v\lab\x) at ($(\z, 0) +(\x*360/8:1)$) {};
    }
    \foreach \x in {0, ..., 7} {
        \pgfmathsetcounter{next}{\x+1}
        \draw (v\lab\x) edge (v\lab\thenext);
    }
}
\node[left] at (0, 0) {$v^*$};
\draw[walk] (vl3) -- (vl4) -- (vl5) -- (vl6) -- (vl7) -- (vl0) -- (vl1);
\end{tikzpicture}
}
}
\caption{Sketch of the graph $G'$. The top shows a configuration where a decision at $v^*$ has to be made. The three configurations below show the possible outcomes. All these transitions happen with equal probability, but only the one on the bottom right leads to a solution where rotations are necessary.}
\label{ds:fig:G'}
\end{figure}
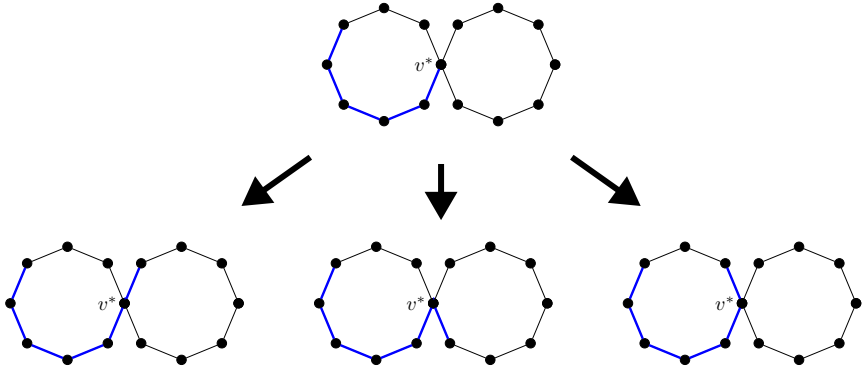

Neumann~\cite{Neumann2008d} considered a simple instance that consists of two cycles of equal size, connected by one common vertex $v^*$ (see Figure~\ref{ds:fig:G'}). The instance is interesting as it represents a worst case for the time until an improvement is found. This is with respect to randomized local search (RLS) working on this representation. RLS works like the \dsea, but it only uses local mutations. As mutation operator it uses jumps: an edge is selected uniformly at random and then it is moved to a (different) target position chosen uniformly at random. All edges in between the two positions are shifted accordingly.

On the considered instance RLS typically starts constructing a walk within one of these cycles, either by appending edges to the end of the walk or by prepending edges to the start of the walk. When the walk extends to $v^*$ for the first time, a decision needs to be made. RLS can either extend the walk to the opposite cycle, see Figure~\ref{ds:fig:G'}. In this case RLS can simply extend both ends of the walk until a Eulerian cycle is formed. The expected time until this happens is $\Theta(m^3)$ where $m$ denote the number of edges.

But if another edge in the same cycle is added at~$v^*$, the walk will evolve into one of the two cycles that make up the instance. It is not possible to add further edges to the current walk, unless the current walk starts and ends in~$v^*$. However, the walk can be rotated so that the start and end vertex of the walk is moved to a neighboring vertex. Such an operation takes expected time $\Theta(m^2)$. Note that the fitness after a rotation is the same as before. Rotations that take the start and end closer to~$v^*$ are as likely as rotations that move it away from~$v^*$. The start and end of the walk hence performs a fair random walk and $\Theta(m^2)$ rotations are needed on average in order to reach~$v^*$. The total expected time for rotating the cycle is hence $\Theta(m^4)$.

Summarizing, if RLS makes the right decision then expected time $\Theta(m^3)$ suffices in total. But if rotations become necessary the expected time increases to $\Theta(m^4)$. Now consider an island model with $\numproc$ islands running RLS. If islands evolve independently for at least~$\tau \ge m^3$ generations, all mentioned decisions are made independently, with high probability. The probability of making a wrong decision is $1/3$, hence with $\numproc$ islands the probability that all islands make the wrong decision is $3^{-\numproc}$, leading to the following result.
\begin{theorem}
\label{ds:the:large-migint-Gprime}
The island model running RLS on $\lambda \le m^{O(1)}$ islands, $\tau \ge m^3$, and an arbitrary topology optimizes $G'$ in expected $O(m^3 + 3^{-\lambda} \cdot m^4)$ generations.
\end{theorem}
The choice $\numproc := \log_3 m$ yields an expectation of $\Theta(m^3)$, and every value up to $\log_3 m$ leads to a superlinear and even exponential speedup, compared to the time $\Theta(m^4)$ for a single island running RLS.

Interestingly, this good performance only holds if migration is used rarely, or if independent runs are used. If migration is used too frequently, the island model rapidly loses diversity. If $T$ is any strongly connected topology and $\diam(T)$ is its diameter, we have the following.
\begin{theorem}
\label{ds:the:lower-bound-unrestricted-Gprime}
Consider the island model with an arbitrary strongly connected topology $T$ running RLS with jumps on each island. If $\migint \cdot \diam(T) \cdot \lambda = O(m^2)$ then the expected number of generations on $G'$ is at least $\Omega(m^4/(\log \lambda))$.
\end{theorem}
If $\tau \cdot \diam(T) \cdot \numproc = O(m^2)$ then there is a constant probability that the island that first arrives at a decision at~$v^*$ propagates this solution throughout the whole island model, before any other island can make an improvement. This results in an expected running time of $\Omega(m^4/\log(\numproc))$. This is almost $\Theta(m^4)$, even for very large numbers of islands. The speedup is therefore logarithmic in the number of islands at best, or even worse.

This natural example shows that the choice of the migration interval can make a difference between exponential and logarithmic speedups.

\section{How Diversity Benefits Crossover}
\label{ds:sec:benefits-for-crossover}

Now we look at examples where diversity enhances the use of crossover in evolutionary algorithms. In a population where all individuals are very similar, crossover is unlikely to be effective as it will create an offspring that is similar to both of its parents. This effect can also be achieved by mutation. Therefore, many examples where crossover is essential require some form of diversity mechanism for crossover to work effectively.

In the following, we review several of these examples, from the very first constructed examples, to problems from combinatorial optimization, and even simple hill climbing problems, where crossover provides a noticeable speedup. In some cases diversity is due to explicit diversity mechanisms; in others, diversity can emerge naturally, from independent variations.

Most of the algorithms discussed in this section fit into the scheme described in Algorithm~\ref{ds:alg:Scheme-GA}.
Unless stated otherwise, parent selection is performed uniformly at random. With a crossover probability $p_c$, crossover is performed; this can be uniform crossover or $k$-point crossover. In any case, mutation is performed with a mutation rate of $p$, which is assumed to be the default value $p=1/n$ unless stated otherwise. In the replacement selection, $\mu$ individuals with the best fitness are selected for survival. In case there are ties, a specific tie-breaking rule can be used; the default is to break ties uniformly at random.

\begin{algorithm2e}
\label{ds:alg:Scheme-GA}
Initialize population $P$ of size $\mu \in \N$ uniformly at random.\\
\While{optimum not found}{%
    Let $P' = \emptyset$.\\
    \For{$i = 1, \dots, \lambda$}{%
        Choose $p \in [0, 1]$ uniformly at random.\\
        \eIf{$p \leq p_c$}
        {
            Select two parents $x_1, x_2$.\\
            Let $y := \text{crossover}(x_1, x_2)$.
        }
        {
            Select a parent $y$.
        }
		Flip each bit in $y$ independently with probability~$p$.\\
        Add $y$ to $P'$.
    }
    \label{ds:line:tie-breaking}Let $P$ contain the $\mu$ best individuals from $P \cup P'$; break ties according to a specified tie-breaking rule.
}
\caption{Scheme of a \dsmlGA with mutation rate~$p$ and crossover with crossover probability~$p_c$ for maximizing $f \colon \{0, 1\}^n \to \R$.}
\end{algorithm2e}

\subsection{Real Royal Road Functions for Crossover}

Jansen and Wegener~\cite{Jansen2005c} were the first to provide an example function for which it could be rigorously proved that a simple genetic algorithm (GA) with crossover takes expected polynomial time, whereas all ($\mu$+$\lambda$)~evolutionary algorithms using only standard bit mutation need exponential time with overwhelming probability.

Their steady-state GA with population size~$\mu$ can be regarded as a special case of Algorithm~\ref{ds:alg:Scheme-GA} with $\lambda=1$, referred to as \dsmuGA in the following, using a tie-breaking rule that eliminates an individual with the largest number of duplicates in $P \cup P'$\footnote{In~\cite{Jansen2005c} the replacement selection stops without altering~$P$ if the fitness of the offspring is smaller than the fitness of the worst individual in~$P$. Our algorithm is equivalent as in this case the offspring will be added and immediately removed from the population.}.
The latter is equivalent to breaking ties towards including individuals with the fewest duplicates in $P \cup P'$.


Jansen and Wegener~\cite{Jansen2005c} define two classes of functions they call \emph{real royal road functions}: one for one-point crossover and one for uniform crossover. We focus on the one for one-point crossover as it is conceptually simpler. Denoting by $b(x)$ the length of the longest block consisting of ones only (e.\,g.\ $b(100${\boldmath$111$}$011) = 3$), the function class $R_n$ is defined as (assuming $n/3 \in \mathbb{N}$)
\[
R_n(x) = \begin{cases}
2n^2 & \text{if $x = 1^n$,}\\
n|x|_1 + b(x) & \text{if $|x|_1 \le 2n/3$,}\\
0 & \text{otherwise.}
\end{cases}
\]
The function contains a strong gradient in the region of search points with at most $2n/3$ ones. The function also contains a fitness valley of fitness 0 that needs to be crossed to reach the optimum~$1^n$. Moreover, the function encourages an evolutionary algorithm to evolve search points with $2n/3$ ones and a maximum block length of $2n/3$. This is to allow crossover to combine two such blocks to create the optimum $1^n$, for instance by crossing over two parents $1^{2n/3}0^{n/3}$ and $0^{n/3}1^{2n/3}$ that have large blocks in different positions.

We give a simplified version of their result as their work includes an additional parameter that specifies the length $m$ of the fitness valley, which is fixed to $m = n/3$ here. Note that the population size~$\mu$ and the crossover probability $p_c$ can be functions of the problem size~$n$.
\begin{theorem}[Simplified from Theorem~3 in~\cite{Jansen2005c}]
Let $p_c \le 1 - \varepsilon$ for some $0 < \varepsilon < 1$, and $\mu \ge 2n/3 + 1$. Then the expected optimization time of the \dsmuGA breaking ties towards including individuals with the fewest duplicates in $P \cup P'$
on $R_n$ is
$
O(\mu n^3 + (\mu \log \mu)n + \mu^2/p_c).
$
For the typical case where $p_c$ is a positive constant and $\mu = O(n)$ the bound is $O(n^4)$.
\end{theorem}
The proof is a beautiful application of the so-called method of typical runs~\cite[Section~11]{Wegener2002}, where a run is divided into phases that reflect the typical behavior of the algorithm. Then the expected times for completing each phase is estimated separately, using arguments most appropriately to that phase. Jansen and Wegener~\cite{Jansen2005c} show that in expected time $O(\mu n)$ the population reaches a state where all search points have $2n/3$ ones (or the optimum is found beforehand). From there, the algorithm can focus on maximizing the maximum block length $b(x)$. In the next expected $O(n^2 \log(n) + (\mu \log \mu)n)$ generations the algorithm evolves a population where all search points~$x$ have the maximum block length $b(x) = 2n/3$ (or the optimum has been found).

Once such a population has been reached, we can rely on the diversity mechanism taking effect: since then all search points have the same fitness, selection for replacement is solely based on the number of duplicates in the population. There are only $2n/3 + 1$ different genotypes with $2n/3$ ones and a block length of $2n/3$:
\[
1^{2n/3}0^{n/3}, \ 01^{2n/3}0^{n/3-1}, \ 0^21^{2n/3}0^{n/3-2}, \ \dots \ , 0^{n/3}1^{2n/3}.
\]
The population size $\mu$ is large enough to be able to store all these search points. Hence, once a particular genotype is created, the population will always retain such a genotype until the optimum is found. Using appropriate 2-bit flips, it is possible to create novel genotypes. The expected time until the population contains all the above genotypes is $O(\mu n^3)$.

Once the population contains the genotypes $1^{2n/3}0^{n/3}$ and $0^{n/3}1^{2n/3}$, if these are selected as parents, one-point crossover can easily create $1^n$ by choosing a cutting point in the middle third of the bit string. The expected time for this event is $O(\mu^2/p_c)$, and summing up all expected times yields the claimed bound.

\subsection{Coloring Problems}
\label{ds:sec:ising}

Fischer and Wegener~\cite{Fischer2005} presented another example where a diversity mechanism enhances crossover for a combinatorial problem. They considered a simple variant of the Ising model, a well-known model of ferromagnetism that is NP-hard to solve in its general case. Here we consider an easy special case where vertices of an undirected graph can have one of two states, 0 and~1 (also referred to as ``colors''), and it is beneficial to color two neighboring vertices with the same color. Then the fitness function corresponds to the number of monochromatic edges, and all colorings where all connected components have the same color are global optima.

The problem is an interesting test bed for evolutionary algorithm because subgraphs of the same color can be regarded as ``building blocks'' of optimal solutions. The inherent symmetry in the problem implies that competing building blocks may emerge, and evolutionary algorithms can get stuck in difficult local optima, depending on the graph.

For bipartite graphs, the problem is equivalent to the well-known \textsc{Graph Coloring} problem, or more specifically, to the 2-coloring problem where the goal is to color the graph with 2 colors such that no two adjacent vertices have the same color, and the fitness function is the number of correctly colored vertices. The reason is that there is a simple bijection between the Ising model and \textsc{Graph Coloring}: flipping all colors of one set of the bipartition turns all monochromatic edges into bichromatic edges, hence turning a solution for the Ising model into a \textsc{Graph Coloring} solution of the same fitness, and vice versa. All results derived for the Ising model variant described above also hold for the 2-coloring problem, if the underlying graph is bipartite.

Fischer and Wegener~\cite{Fischer2005} studied ring graphs (or cycle graphs) where the $i$-th vertex is neighbored to vertices $i-1$ and $i+1$ (identifying vertex 0 with vertex $n$), and observed that the fitness landscape contains lots of plateaus. A search point such as $0001111000$ contains blocks of bits with the same color, e.\,g., a block of four 1-bits. A mutation flipping only the first or the last bit of such a block can shorten the block; a mutation flipping only the last bit before the block or the first bit following the block can enlarge it. Those mutations are fitness-neutral (i.\,e.\ do not change the fitness), unless some block disappears, which leads to an increase in fitness. The main results from~\cite{Fischer2005} are as follows. Note that rings with an even number of vertices are bipartite, allowing us to add statements about 2-coloring.

\begin{theorem}[Adapting Theorems~4 and~5 in~\cite{Fischer2005}]
The expected optimization time for the \dsea on the Ising model and the 2-coloring problem on rings with even~$n$ is $O(n^3)$.
This bound is asymptotically tight when starting with two blocks of length $\varepsilon n$ and $(1 - \varepsilon)n$, $0 < \varepsilon < 1/2$
a constant.
\end{theorem}
The main observation is that the length of any block follows a fair random walk (apart from boundary states), with a large self-loop probability as the probability of changing the length of the considered block is $\Theta(1/n)$. From the described setting with two blocks, it takes $\Theta(n^2)$ of these changes for some block to disappear, which results in a global optimum.

Using a simple GA with fitness sharing, however, is able to find a global optimum in expected time $O(n^2)$.
\begin{theorem}
\label{ds:the:ising-on-ring}
Consider a (2+2)~GA as a variant of Algorithm~\ref{ds:alg:Scheme-GA} that with probability $p_c = 1/2$ applies two-point crossover once to create two offspring and uses fitness sharing with sharing radius $\sigma = n$ to select 2 search points amongst parents and offspring that maximizes the shared fitness of the population.
The expected number of fitness evaluations until this GA
finds an optimum for the Ising model and the 2-coloring problem on rings with even~$n$ is bounded by $O(n^2)$.
\end{theorem}
The main observation is that fitness sharing turns plateaus into gradients as it rewards the creation of dissimilar individuals. The GA then efficiently creates two complementary individuals (e.\,g.\ through 1-bit flips), and then two-point crossover is able to invert whole blocks, provided that the cutting points are chosen between two blocks, e.\,g.\ turning $00\boldsymbol{11001}0$ into $00001100$ by replacing the bits in bold with the values from the complementary parent.

A similar, but more drastic effect was also shown for coloring complete binary trees~\cite{Sudholt2005}. Here subtrees of the same color represent building blocks of good solutions. The problem is much harder than coloring rings as it contains difficult local optima, for example when the two subtrees of the root are colored with different colors.

This author showed that all algorithms in a large class of \dsmlea{}s, with arbitrary mutation rates, need at least expected time $2^{\Omega(n)}$ to find a global optimum. In contrast, the (2+2)~GA with fitness sharing finds an optimum in expected polynomial time.
\begin{theorem}
Consider the (2+2)~GA with fitness sharing described in Theorem~\ref{ds:the:ising-on-ring}. The expected optimization time for the Ising model and the 2-coloring problem on a complete binary tree with $n$ vertices is bounded by $O(n^3)$.
\end{theorem}
The analysis shows that fitness sharing again encourages an increase in the Hamming distance between the two current search points, $x$ and $y$.
In the case of $x$ and $y$ being complementary, two-point crossover can effectively substitute subtrees to increase the fitness. The challenge lies in showing that and how complementary search points evolve.
In contrast to rings, binary trees do not contain any plateaus, hence it is not always possible to increase the Hamming distance without compromising on fitness. Interestingly, a case distinction according to the function $H(x, y)+f(x)+f(y)$ shows that if this function is small then there are accepted 1-bit flips that increase the real fitness, possibly at the expense of decreasing the Hamming distance $H(x, y)$. But if the function is large, there are accepted 1-bit flips that increase the Hamming distance at the expense of the real fitness. There is a ``gray area'' in between, where more complex operations (mutation and/or crossover) are required; however, these steps have probability $\Omega(1/n^2)$, leading to the overall time bound of $O(n^3)$.

In this scenario, even though fitness sharing can maximize diversity at the expense of the real fitness, it turns out to be an effective strategy, as the diversity of complementary search points can be exploited efficiently by crossover.

\subsection{Diversity and Crossover Speed Up Hill Climbing}
\label{ds:sec:speed-up-hill-climbing}


Diversity and crossover also prove useful in a very natural and well-known setting, albeit with smaller speedups compared to the examples seen so far.
The simple problem \textsc{OneMax} is the most-studied problem in theory of randomised search heuristics. It can be regarded as a simple hill climbing task, as a mutation flipping a single 0-bit to 1 increases the fitness. It can also be seen as a problem where ones are ``building blocks'' of the global optimum, and the algorithm has to assemble all building blocks to find the optimum. This perspective is related to the so-called  ``building block hypothesis'', an attempt to explain the advantage of crossover as GAs with crossover can combine building blocks of good solutions. Yet, it has been surprisingly hard to come up with natural examples and rigorous proofs to cement or refute this hypothesis.

This author~\cite{Sudholt2016,Sudholt2012b} showed that the \dsmlGA with the duplicate-based tie-breaking rule is twice as fast as the fastest evolutionary algorithm using only standard bit mutation (modulo small-order terms).
\begin{theorem}[Simplified from Theorem~1 and Theorem~4 in~\cite{Sudholt2016}\footnote{We remark that results in~\cite{Sudholt2016} hold for much larger ranges of the mutation rate~$p$ and arbitrary parent selection mechanisms that do not disadvantage individuals with higher fitness.}]
\label{ds:the:upper-bound-uniform-crossover-variable-p-every-GA}
Let $n \ge 2$ and $c > 0$ be a constant.
Every evolutionary algorithm that uses only standard bit mutation with mutation rate~$p=c/n$ to create new solutions has expected optimization time at least
$
\frac{e^{c}}{c} \cdot n \ln n \cdot (1-o(1))
$
on \onemax and every other function with a unique optimum.

The expected optimization time of the \dsmlGA breaking ties towards including individuals with the fewest duplicates in $P \cup P'$, with $0 < p_c < 1$ constant, mutation probability~$p=c/n$ and $\mu, \lambda = o((\log n)/(\log \log n))$, on \onemax is at most
$
\label{ds:eq:upper-bound-GAs-one-plus-o-one}
\frac{e^c}{c \cdot (1+c)} \cdot n \ln n \cdot (1 + o(1)).
$
\end{theorem}
Modulo small-order terms, this is a speedup of $1+c$, which is $2$ for $c=1$, reflecting the default mutation rate $p=1/n$.

The idea behind the proof is to make a case distinction for all possible populations, according to the current best-so-far fitness and the diversity in the population, and then to upper bound the expected time spent in all these cases.

If a population contains individuals of different fitness values, the individuals of current best-so-far fitness~$i$ quickly take over the population (or an improvement of the best-so-far fitness is found). Due to our restrictions on population sizes $\mu, \lambda$, the total time across all best-so-far fitness values is $o(n \log n)$. If the population consists of $\mu$ identical genotypes of fitness~$i$, this state will be left for good if either a fitness improvement is found, or a different individual with the same fitness is created. In the latter case, the diversity mechanism in the tie-breaking rule ensures that this diversity will never get lost (unless an improvement is found).

The mentioned diversity can be created by a fitness-neutral mutation that flips the same number of 0-bits to 1 as it flips 1-bits to 0.
Such a multi-bit flip would be irrelevant for mutation-only evolutionary algorithms. But when crossover is used, it can exploit the diversity created this way by choosing two parents with equal fitness but different genotypes, and to create a surplus of ones on the bit positions where the two parents differ. Creating such a surplus is very likely; the probability for such an event is at least~$1/4$, irrespective of the Hamming distance between the two parents. The time the algorithm spends evolving a diverse population is negligible compared to the time spent in a state where all individuals are identical.

The expected time is thus dominated by the time spent trying to leave states where all genotypes are identical. Compared to mutation-based evolutionary algorithms, the creation of diversity offers another route towards fitness improvements as crossover rapidly exploits this diversity, creating improvements almost instantly.

Corus and Oliveto~\cite{Corus2017b} recently showed that the choice of tie-breaking rule is important to get the above-mentioned speedup: when replaced with a uniform tie-breaking rule, we still get a constant-factor speedup, but the constant is worse.
\begin{theorem}[Simplified from Theorem~9 in~\cite{Corus2017b}]
The expected optimization time of the \dsmuGA with uniform tie-breaking, ${p_c=1}$, mutation probability~$p=c/n$ and $3 \le \mu = o((\log n)/(\log \log n))$, on \onemax is at most
$
\label{ds:eq:upper-bound-GAs-one-plus-o-one}
\frac{e^c}{c \cdot (1+c/3)} \cdot n \ln n \cdot (1 + o(1)).
$
\end{theorem}
For the standard mutation rate of~$1/n$, the previous speedup of~2 from Theorem~\ref{ds:the:upper-bound-uniform-crossover-variable-p-every-GA} now becomes a factor of~$4/3$. This constant is best possible under mild assumptions~\cite[Theorem~11]{Corus2017b}.


\subsection{Overcoming Fitness Valleys with Naturally Emerging Diversity and Crossover}
\label{ds:sec:jump-emerging}

In Section~\ref{ds:sec:speed-up-hill-climbing} we have seen that diversity and crossover can speed up hill climbing on \onemax by a constant factor. Now we consider the task of overcoming fitness valleys in order to solve multimodal problems. We specifically focus on the problem class $\mathrm{Jump}_k$, the first example function where crossover was proven to be beneficial~\cite{Jansen2002}:
 \[
    \mathrm{Jump}_k(x) =
    \begin{cases}
        k + |x|_1  &\textrm{if } |x|_1 = n \textrm{ or } |x|_1 \leq n - k,\\
        n - |x|_1	&\textrm{otherwise,}
    \end{cases}
\]
In this problem, GAs have to overcome a fitness valley such that all local optima have $n-k$ ones and thus Hamming distance~$k$ to the global optimum~$1^n$. Jansen and Wegener~\cite{Jansen2002} showed that, while mutation-only algorithms such as the \dsea require expected time $\Theta(n^k)$,  a simple \dsmuGA with crossover only needs time $O(\mu n^2 k^3 + 4^k/p_c)$. This time is $O(4^k/p_c)$ for large~$k$, and hence significantly faster than mutation-only GAs.

The factor $4^k/p_c$ results from the fact that, if the population contains pairs of parents that do not share a common 0-bit, then uniform crossover can set all the $k+k=2k$ bits where exactly one parent has a 1 to 1 in the offspring, with probability $2^{-2k} = 4^{-k}$. Hence the expected time for a successful crossover that creates the optimum is bounded by $4^k/p_c$.
Note that such two parents have the largest possible Hamming distance~$2k$ between local optima, hence populations typically achieve a maximum possible diversity between many pairs of parents.
A drawback of their analysis is that it requires an unrealistically small crossover probability $p_c \le 1/(ckn)$ for a large constant~$c > 0$.

K{\"o}tzing, Sudholt, and Theile~\cite{Koetzing2011a} later refined these results towards a crossover probability $p_c \le k/n$, which is still unrealistically small.
Both approaches focus on creating a maximum Hamming distance between local optima through a sequence of lucky mutations, relying on crossover to create the optimum, once sufficient diversity has been created. Their arguments break down if crossover is applied frequently. Hence, these analyses do not reflect the typical behaviour in GA populations with constant crossover probabilities $p_c = \Theta(1)$ as used in practice.

We review recent results from Dang \emph{et al.}~\cite{Dang2017} where realistic crossover probabilities were considered, at the expense of a smaller (but still significant) speedup.
Previous work~\cite{Jansen2002,Koetzing2011a} relied on independent mutations providing diversity, and regarded crossover as potentially harmful, as the effect of crossover on diversity was not well understood. This led to a worst-case perspective on crossover: previous proofs considered mutation to build up diversity over time, like a house of cards, with the worst-case assumption that one unexpected application of crossover would destroy the build-up of diversity, collapsing the house of cards, and the build-up of diversity had to restart from scratch. This view is backed up by a negative result~\cite[Theorem~8]{Koetzing2011a}, showing that if using only crossover with $p_c = \Omega(1)$ but no mutation following crossover, diversity reduces quickly, leading to inefficient running times for small population sizes ($\mu = O(\log n)$).

In~\cite{Dang2017} a different perspective was offered, an approach loosely inspired from population genetics: the authors showed that crossover, when followed by mutation, can actually be very beneficial in creating diversity. Note that the perspective of crossover creating diversity is common in population genetics~\cite{Komarova,Weissman1389}. A frequent assumption is that crossover mixes all alleles in a population, leading to a situation called \emph{linkage equilibrium}, where the state of a population is described by the frequency of alleles~\cite{Barton:2013:QPG:2463372.2463568}.

The main result can be stated as follows.
\begin{theorem}[Theorem~6 in~\cite{Dang2017}, simplified for $k \ge 3$]
\label{ds:the:upper-bound-no-mechanism}
The expected optimization time of the \dsmuGA with $p_c=1$ and $\mu \le \kappa n$, for some constant~$\kappa > 0$, on $\mathrm{Jump}_k$, $3 \le k = o(n)$, is
$
O(n^k/\mu + n^{k-1}\log(\mu))
$.
\end{theorem}
For $\mu = \kappa n$, the bound simplifies to $O(n^{k-1} \log n)$, a speedup of order $\Omega(n/\!\log n)$ compared to the expected time of $\Theta(n^k)$ for the \dsea~\cite{Jansen2002}.

The analysis shows that on
$\mathrm{Jump}_k$ diversity emerges naturally in a population: the interplay
of crossover, followed by mutation, can serve as a catalyst for
creating a diverse range of search points out of few different
individuals. Consider the situation where all individuals in the population are local optima with $n-k$ ones, and assume pessimistically that there is no diversity: all individuals are identical. In the following we refer to a collection of identical individuals with $n-k$ ones as a \emph{species}. Mutation is able to create a new species, for instance by flipping a single 0-bit and a 1-bit. This new species can grow in size, or become extinct over time.

Crossing over two individuals from different species can easily create a surplus of ones, where the offspring has $n-k+1$ ones. The following mutation now creates a local optimum if it flips a 1-bit back to 0. Note that here there are $n-k+1$ ones to choose from, each leading to a different species. This means that, once mutation has created a small amount of diversity, crossover and mutation can work together in this way to create a burst of diversity that has a good chance to prevail for a long time, before the population loses all diversity, or the global optimum is found.

In the proof of Theorem~\ref{ds:the:upper-bound-no-mechanism}, the size of the largest species is taken as a potential function: if the size of the largest species is $\mu$, there is no diversity, but if it is bounded away from~$\mu$, it is easy to select two parents from different species with uniform parent selection. The size of the largest species behaves like an almost fair random walk, and the population has a good chance of spending long periods of time in states where the size of the largest species is small. In these situations, when two parents from different species are selected, crossover has a chance to create a surplus of 1-bits, and then the global optimum can be found by flipping the at most $k-1$ remaining 0-bits to~1.

This argument also shows that speedups can be achieved from small amounts of diversity; in contrast to previous work~\cite{Jansen2002,Koetzing2011a} it is not necessary to rely on a maximum Hamming distance between parents emerging.

A further finding from~\cite{Dang2017} is that increasing the mutation rate to
$p=(1+\delta)/n$ for an arbitrarily small constant $\delta>0$ turns the almost fair random walk describing the size of the largest species into an
unfair random walk that is biased towards increased diversity. In other words, larger mutation rates facilitate the emergence and maintenance of diversity in this setting. This leads to the following improved upper bound, which for reasonably small~$\mu$ gives a speedup of order~$n$ over the expected time of the \dsea.
\begin{theorem}[Theorem~10 in~\cite{Dang2017}, simplified for $k \ge 3$]\label{ds:thm:highmutationruntime}
  The \dsmuGA with mutation rate $(1+\delta)/n$, for a constant
  $\delta>0$, and population size $\mu\geq ck\ln(n)$ for a
  sufficiently large constant $c>0$, has  for $3 \le k=o(n)$ expected optimization time
  $O(\mu^2+n^{k-1})$ on $\mathrm{Jump}_k$.
\end{theorem}

\subsection{Speeding Up Fitness Valley Crossing with Explicit Diversity Mechanisms}
\label{ds:sec:jump-diversity-mechanisms}


The performance of the \dsmuGA on $\mathrm{Jump}_k$ can be further improved by using explicit diversity mechanisms in the tie-breaking rule of the \dsmuGA. This was studied in Dang \emph{et al.}~\cite{Dang2016}, where the main results are summarized in Table~\ref{ds:tab:results-diversity-on-jump}. For comparison, the table also contains results reviewed in Section~\ref{ds:sec:jump-emerging} for uniform tie-breaking, where no diversity mechanism is used.

\begin{table*}[tb]
\centering
    \begin{tabular}{l@{\ \ }l@{\ \ }l}
        \hline
        \textbf{Mechanism} & \textbf{General~$\boldsymbol{\mu}$, $\boldsymbol{p_c}$} & \textbf{Best~$\boldsymbol{\mu}$, $\boldsymbol{p_c}$}\\
        \hline
        {None, $p=1/n$} & $O(n^k/\mu + n^{k-1}\log \mu)$ & $O(n^{k-1}\log n)$\\
        {None, $p=(1+\delta)/n$} & $O(\mu^2 + n^{k-1})$ & $O(n^{k-1})$\\
        {Duplicate elimination} & $O(\mu^2 n + n^{k-1})$ & $O(n^{k - 1})$\\
        {Duplicate minimization} & $O(\mu n + n^{k-1})$ & $O(n^{k - 1})$\\
        {Deterministic crowding} & $O(\mu n + n \log n + n e^{5k} \mu^{k+2})$ & $O(n\log n + ne^{5k}2^{k})$\\
        {Convex hull max.} & $O(\mu n^2\log n + 4^k/p_c)$ & $O(n^2\log n + 4^k)$\\
        {Hamming distance max.} & $O(n \log n + \mu^2kn \log(\mu k) + 4^k/p_c)$ & $O(n\log n + nk\log k + 4^k)$\\
        {Fitness sharing} & $O(n \log n + \mu^2 kn \log(\mu k))$ & $O(n\log n + nk\log k + 4^k)$\\
        {Island model} & $O(n \log n + \mu^2 kn + \mu^2 4^k)$ & $O(n\log n + kn + 4^k)$\\
        \hline
    \end{tabular}
    \caption{\label{ds:tab:results-diversity-on-jump} Overview over the main results from~\cite{Dang2016} and~\cite{Dang2017} (Theorems~\ref{ds:the:upper-bound-no-mechanism} and~\ref{ds:thm:highmutationruntime}), restricted to $3 \le k = o(n)$. Most bounds come with mild restrictions on $k, \mu$, or $p_c$, see~\cite{Dang2016,Dang2017} for details. The second column shows simplified bounds, assuming a choice of $\mu$ and $p_c$ that yields the best possible upper bound.
    }
\end{table*}

The different mechanisms (except for the island model) only appear in the tie-breaking rule; they are described as follows, along with the main ideas behind their analysis.
\begin{description}
\item[\bf Duplicate elimination] always chooses an individual for removal that has duplicates in the population, if duplicates exist. Otherwise, it removes an individual uniformly at random. The analysis shows that after $O(\mu^2 n)$ generations in expectation, there will only be $(1-\Omega(1))\mu$ duplicates in the population, and this property will be maintained forever. Then the probability of picking non-identical parents is $\Omega(1)$.
    Then, as argued in Section~\ref{ds:sec:jump-emerging}, crossover followed by mutation can find the optimum with probability $\Omega(n^{k-1})$ as crossover creates a surplus of 1-bits with probability $\Omega(1)$ and then mutation has to flip at most $k-1$ bits to reach the optimum.
\item[\bf Duplicate minimization] is the familiar rule that breaks ties towards including individuals with the fewest duplicates in $P \cup P'$. Here it is easy to show that the size of the largest species decreases to $(1-\Omega(1))\mu$ in expected time $O(\mu n)$. Then we apply the trail of thought from duplicate elimination.
\item[\bf Deterministic crowding] in the case of fitness ties always removes the parent, if the offspring was created by mutation only, or one of the two parents chosen uniformly at random, if the offspring was created by crossover and mutation. The analysis follows the approach from~\cite{Koetzing2011a}, relying on a sequence of events that evolves in a pair of search points that have a maximum Hamming distance of~$2k$. Then there is a reasonable chance that uniform crossover will create the optimum from crossing over these parents and setting all differing bits to~1.
\item[\bf Convex hull maximization] breaks ties towards maximizing the convex hull of the population, which is the set of search points that can be produced from uniform crossover of any two parents. More precisely, we maximize the convex hull by maximizing the number of bit positions where the population contains both a 0 and a 1 in some individual. The analysis shows, similar to~\cite{Gao2014}, that in expected time $O(\mu n^2 \log n)$ a maximum amount of diversity is created, where all of the $\mu k \le n$ zeros in the population occupy different bit positions. Then any two (different) parents have maximum Hamming distance~$2k$ and the optimum can be constructed with probability $p_c \cdot 4^{-k}$ (as argued earlier).
\item[\bf Hamming distance maximization] breaks ties towards maximizing the total Hamming distance between all pairs of search points. Similar to convex hull maximization, we reach a population of maximal diversity in expected time $O(\mu^2 kn \log(\mu k))$. Repeating the arguments from there yields the claimed bound.
\item[\bf Fitness sharing] with a sharing radius of $\delta \ge 2k$ in the setting of populations with equal fitness turns out to be equivalent to maximizing the total Hamming distance between pairs of search points, hence the previous analysis carries over.
\item[\bf Island model] uses a particular topology called \emph{single-receiver model}~\cite{Watson2007}, where $\mu$ islands run a (1+1)~EA independently, and there is a single receiver island that in every generation chooses to islands uniformly at random, copies their current search points, and performs a uniform crossover on these. The analysis shows that, when fixing any two islands, these islands will either have zeros in different positions, or they will, in expectation, reduce the number of bit positions where they have a zero in common. Once the islands have no zero in common, the receiver island has a good chance to create the optimum when crossing over individuals from these two islands.
\end{description}


The island model with single-receiver topology was introduced in~\cite{Watson2007} where the authors relied on this diversity mechanism to prove that a constructed royal road function with a building block structure could be solved efficiently by crossover. It was also used in~\cite{Neumann2011} where it was shown that crossover during migration can be effective, for constructed functions as well as for instances of the \textsc{Vertex Cover} problem. We refer the reader to~\cite{Neumann2011,Watson2007} and the survey~\cite[Section~46.5.4]{Sudholt2012a} for details.

\section{How Diversity Benefits Dynamic Optimisation}
\label{ds:sec:dynamic}

Another very important use of population diversity is to ensure good performance in dynamic optimization, where the problem can change over time. Diversity can ensure that the population is able to keep track of global optima, or to re-discover global optima in case different local optima change their fitness, and another local optimum becomes the new global optimum.

The runtime analysis of dynamic evolutionary optimization is still in its very infancy, with only few results available (e.\,g.~\cite{Dang2017a,Jansen2014a,Kotzing2015,Koetzing2012b,Lissovoi2015,Oliveto2015,RohlfshagenLehreYao2009}).

\subsection{Diversity Mechanisms for Balance}

Oliveto and Zarges~\cite{Oliveto2015} considered diversity mechanisms for the dynamic function \textsc{Balance}~\cite{RohlfshagenLehreYao2009}.

\begin{definition}[\textsc{Balance}~\cite{RohlfshagenLehreYao2009}]\label{ds:def-balance}
Let $a,b \in \{0,1\}^{n/2}$ and $x = ab \in \{0,1\}^n$. Then
\begin{equation*}
\textsc{Balance}(x) =
	\begin{cases}
		n^3 & \text{if } \textsc{LO}(a) = n/2, \text{else}\\
		|b|_1 + n \cdot \textsc{LO}(a) & \text{if } n/16 < |b|_1 < 7n/16 , \text{else}\\
		n^2 \cdot \textsc{LO}(a) & \text{if }  |a|_0 > \sqrt{n}, \text{else}\\
		0 & \text{otherwise}
	\end{cases}
\end{equation*}
where $|x|_1 = \sum_{i=1}^{n/2} x_i$,  $|x|_0 = n/2 - |x|_1$ is a number of zeros and $\textsc{LO}(x) :=\sum_{i=1}^{n/2} \prod_{j=1}^i x_j$ counts the number of leading ones.
\end{definition}
For the majority of search points, the function gives hints to maximize the number of ones in the suffix (also referred to as \onemax part), and even stronger hints to maximize the number of leading ones in the prefix (the leading ones part). All search points with a maximum of $n/2$ leading ones are global optima, however the function also contains two traps and a fitness valley of fitness 0 that separates the traps from the region of global optima. The upper trap contains all search points with more than $7n/16$ ones, and the lower trap contains all search points with less than $n/16$ ones.

The function is used in a dynamic framework where every $\tau$ generations, for a change frequency parameter~$\tau$, the roles of zeros and ones in the suffix is reversed, so that the fitness gradient switches between maximizing and minimizing the number of ones in the \onemax part. Unless stated otherwise, the arguments given below assume that the number of ones is maximized.

Oliveto and Zarges~\cite{Oliveto2015} show that a \dsmuea with no diversity mechanism tends to fail on \textsc{Balance}, as the whole population is likely to run into one of the traps.
\begin{theorem}
\label{ds:the:dynamic-no-mechanism}
If $\tau>20 \mu n$ and $\mu \le n^{1/2-\varepsilon}$ then expected time for the \dsmuea to optimize \textsc{Balance} is at least
$n^{\Omega(\sqrt{n})}$. If $\tau > 38\mu n^{3/2}$ and $\mu \le n^{1/2-\varepsilon}$ then the \dsmuea requires at least $n^{\Omega(\sqrt{n})}$ steps with overwhelming probability.
\end{theorem}
The intuitive reason for this poor performance is that it is easier to optimize \onemax than it is to maximize the number of leading ones, and the algorithm only needs to come moderately close to the \onemax (or \textsc{ZeroMax}) optimum to fall into a trap. With low frequencies of change, this is very likely to happen.

The authors investigated in how far this poor performance can be mitigated by using diversity-preserving mechanisms like the ones studied for \twomax in~\cite{Friedrich2009}. The main results are explained in the following.

Genotype diversity, that is, preventing genotype duplicates from being accepted, is too weak to affect the main search behavior; the \dsmuea still tends to run into traps.
\begin{theorem}
For the \dsmuea with genotype diversity (Algorithm~\ref{ds:alg:muea-avoiding-genotypes}), the results from Theorem~\ref{ds:the:dynamic-no-mechanism} apply.
\end{theorem}

Deterministic crowding does not help: recall that deterministic crowding is based on offspring competing against their direct parents, hence (since no crossover is used) the \dsmuea evolves $\mu$ independent lineages. Each lineage still has a high probability of running into a trap, hence for polynomial population sizes and low frequencies of change there is a high probability that the whole population will be led into a trap.
\begin{theorem}
With overwhelming probability, the \dsmuea using deterministic crowding and $\mu \le n^{O(1)}$ requires exponential time to optimize \textsc{Balance} if $\tau > 8e\mu n$.
\end{theorem}

Fitness diversity as in Algorithm~\ref{ds:alg:muea-fitness-diversity} turns out to perform a lot better: it can find the optimum efficiently for all values of $\tau$. This is surprising as this mechanism showed the worst performance for \textsc{TwoMax}~\cite{Friedrich2009}.
\begin{theorem}
Let $\mu > n - 2(\sqrt{n} - 1)$. Then with overwhelming probability, the \dsmuea with fitness diversity optimizes \textsc{Balance} in time $O(\mu n^3)$ for arbitrary $\tau \ge 0$.
\end{theorem}
The proof shows that, as the population size is quite large, the \dsmuea is able to ``fill up'' both traps in a sense that the algorithm will eventually contain individuals representing all fitness values inside a trap, and then no other point in the trap will be accepted. This then allows the algorithm to evolve a lineage leading to the global optimum, avoiding the trap.

Finally, the authors consider a variant of the \dsmuea with population size~$\mu=2$ combining fitness sharing (with $\alpha=1$ and sharing radius $\sigma=n$) and deterministic crowding: in the selection step the shared fitness of the current population is compared against the shared fitness of the population where the offspring replaces its parent, and the latter population is selected if its shared fitness is no smaller. Instead of standard bit mutation, local mutations are used that flip exactly one bit chosen uniformly at random, as done in RLS. The resulting algorithm is referred to as \dstworls.

The \dstworls is efficient with probability close to~$1/2$.
\begin{theorem}
With probability at least $1/2 - e^{-\Omega(n)}$ the \dstworls with fitness sharing and crowding finds the optimum of \textsc{Balance} in time $O(n^2)$ for arbitrary $\tau \ge 0$.
\end{theorem}
The analysis observes, similar to~\cite{Sudholt2005} reviewed earlier in Section~\ref{ds:sec:ising}, that the function $H(x, y) + f(x) + f(y)$ for current search points $x$, $y$ decides whether $H(x, y)$ can be increased at the expense of fitness, or whether the fitness can increase at the expense of the Hamming distance. Bit flips in the \onemax part of \textsc{Balance} only change the fitness by~1. If $H(x, y) + f(x) + f(y) > 2n$, such bit flips are accepted if and only if they increase the Hamming distance. If $H(x, y) + f(x) + f(y) < 2n$, such bit flips are accepted if and only if they increase the fitness.

Now, a fitness larger than $2n$ is easily achieved if at initialization $x$ and $y$ have a total of at least 2 leading ones. This happens with probability at least $1/2$, and then the \dstworls will always have a fitness larger than~$2n$. Then any bit flips in the \onemax part will only be accepted if they increase the Hamming distance~$H(x, y)$. With high probability there will be many bit positions~$i$ where $x_i = 1, y_i = 0$ and many bit positions~$j$ where $x_j = 0, y_j = 1$. These values will never change, hence the \onemax part of any search point will never meet the extreme values corresponding to a trap. The leading ones part will be optimized as it has a much larger impact on the fitness, leading to a global optimum in the claimed time.

However, the algorithm can also fail badly with constant probability, getting stuck in a local optimum from which there is no escape.
\begin{theorem}
Let $\tau >12n +1$. With probability bounded below by a constant the \dstworls with fitness sharing and crowding requires infinite time to optimize \textsc{Balance}.
\end{theorem}
This statement can be shown by observing that with constant probability, the fitness will remain below $2n$, hence maximizing the number of ones in the \onemax part, while one of the search points reaches the upper trap. Then the fitness will always be larger than $2n$, which makes the algorithm maximize the Hamming distance and hence drives the other search point into the lower trap. Here the algorithm gets stuck as the traps cannot be left and local mutations cannot create the global optimum from a trap.

An interesting conclusion when contrasting the performance of diversity mechanisms on \twomax~\cite{Friedrich2009} and \textsc{Balance}~\cite{Oliveto2015} is that mechanisms that perform well on one function may not perform well on the other. Fitness diversity shows the worst performance guarantees for \twomax, but it performs the best on \textsc{Balance}. Deterministic crowding performed well on \twomax, but performs poorly on \textsc{Balance}.
Fitness sharing performed the best on \twomax, but is only effective on \textsc{Balance} with constant probability, and otherwise fails badly.

\subsection{Island Models for the Maze Function}

Lissovoi and Witt~\cite{Lissovoi2017a} presented another example where diversity mechanisms prove useful in dynamic optimization. They showed that island models can help to optimize the dynamic function \textsc{Maze}, introduced earlier by K{\"o}tzing and Molter~\cite{Koetzing2012b}. The function \textsc{Maze} changes in phases of $t_0$ steps. In the first phase, the function is equivalent to \textsc{OneMax}. In the next $n$ phases, higher fitness values are assigned to two search points on a shortest Hamming path from $1^n$ to $0^n$, in an oscillating pattern. Every two iterations out of three, $0^i1^{n-i}$ receives fitness $n+2$, while the previous point on the path, $0^{i-1}1^{n-i+1}$, receives fitness $n+1$. Every three iterations, the fitness values of these two points are reversed. In every phase the index~$i$ increases by~1. All other search points always retain their \textsc{OneMax} value, hence whenever an algorithm loses track of the path, it is likely to be led back into~$1^n$. The optimum can only be reached if an algorithm tracks the moving optimum on the whole Hamming path, eventually reaching $0^n$ after $n$ phases. We refer to~\cite{Koetzing2012b,Lissovoi2017a} for formal definitions of \textsc{Maze}.

The \dsea fails badly on \textsc{Maze}~\cite{Koetzing2012b} and the same holds for a (1+$\lambda$)~EA with a moderate offspring population size, as shown in the following theorem. The reason is that in every phase there is a constant probability that the algorithm will maintain the previous point on the path, $0^{i-1}1^{n-i+1}$, and will fall off the path once the next phase starts.
\begin{theorem}
The (1+$\lambda$)~EA with $\lambda = O(n^{1-\varepsilon})$, for any constant $\varepsilon > 0$, will with
high probability lose track of the optimum of \textsc{Maze}, i.\,e., with high probability it will require an exponential number of iterations to construct the final optimum.
\end{theorem}

In sharp contrast, a simple island model running $\lambda$ \dsea{}s is effective on \textsc{Maze}, even with a much smaller number $\lambda$ of offspring created in each generation.
\begin{theorem}
An island model with $\lambda = c \log n$ islands, where $c$ is a sufficiently large constant, each island running a \dsea, and migration along a complete topology occurring during the first iteration of every phase (i.\ e., with migration interval $\tau = t_0$) is able to find the optimum of the \textsc{Maze} with phase length $t_0 = kn^3 \log n$ in polynomial time with high
probability.
\end{theorem}
The intuitive reason is that each island on the path has a constant probability of ending the phase in $0^{i-1}1^{n-i+1}$ and a constant probability of ending it in $0^{i}1^{n-i}$. In the latter case, these islands will still be on the path once the index~$i$ increases at the start of the next phase. There is a high probability that at least one island will still be on the path, and its fitness will be no less than that of all the other islands. Hence migration will ensure that all islands that may have fallen off will be put back on to the path.

Note that the choice of the migration interval aligns with the time interval $t_0$ for dynamic changes, such that at the time of migration, the search points further up on the path have a higher fitness. If migration occurs at other points in time, island models with $O(\log n)$ islands may still fail on \textsc{Maze}~\cite[Theorem~14]{Lissovoi2017a}.

\section{Diversity-Based Parent Selection}
\label{ds:sec:diversity-based-parent-selection}

All results surveyed so far use diversity mechanisms in the environmental selection, i.\,e.\ to decide which search points are allowed to survive to the next generation. Here we present recent work by Covantes Osuna et al.~\cite{CovantesOsuna2017a}, who suggested to use diversity mechanisms in the parent selection in the context of evolutionary multiobjective optimization.

Well established multi-objective evolutionary algorithms (MOEAs) such as NSGA-II~\cite{Deb2002}, SPEA2~\cite{Bleuler2001}, IBEA~\cite{Zitzler2004} have two basic principles driven by selection. First of all, the goal is to push the current population close to the ``true'' Pareto front. The second goal is to ``spread'' the population along the front such that it is well covered. The first goal is usually achieved by dominance mechanisms between the search points or indicator functions that prefer non-dominated points. The second goal involves the use of diversity mechanisms. Alternatively, indicators such as the hypervolume indicator play a crucial role to obtain a good spread of the different solutions of the population along the Pareto front.

In the context of EMO, parent selection is usually uniform whereas offspring selection is based on dominance and the contribution of an individual to the diversity of the population. The work~\cite{CovantesOsuna2017a} shows that diversity mechanisms can also be highly beneficial when embedded into the parent selection mechanisms in EMO. The goal is to speed up the optimization process of an EMO algorithm by selecting individuals that have a high chance of producing beneficial offspring.
The idea is to use a diversity metric, such as the hypervolume contribution or the crowding distance contribution, and to preferably select parents with a higher diversity score. The hypervolume describes the area that is dominated by points in the population; the hypervolume contribution describes the contribution a search point~$x$ makes to the hypervolume, i.\,e.\ the difference between the hypervolume of the whole population~$P$ and the population $P \setminus \{x\}$ without~$x$. The crowding distance is a well-known measure from NSGA-II; it is based on the distances in objective space to the search points with the closest objective values, considering each objective separately.

The main assumption is that individuals with a high diversity score are located in poorly explored or a less dense areas of the search space, so the chances of creating new non-dominated individuals are better than in areas where there are several individuals. In this sense the new parent selection schemes focus on individuals where the neighbourhood is not fully covered and in consequence, force the reproduction in those areas and to the spread of the population along the search space.

We consider two well-known pseudo-Boolean functions $\{0,1\}^n\to \mathbb{N}^2$ with two objectives: in
\begin{displaymath}
\textsc{OneMinMax}(x_1,\ldots,x_n):= \left(n-\sum_{i=1}^{n}x_i,\sum_{i=1}^{n}x_i\right),
\end{displaymath}
the aim is to maximize the number of zeroes and ones at the same time.
For
\begin{displaymath}
\textsc{LOTZ}(x_1,\ldots,x_n):=\left(\sum_{i=1}^{n}\prod_{j=1}^{i}x_j,\sum_{i=1}^{n}\prod_{j=i}^{n}(1-x_j) \right),
\end{displaymath}
the goal is to simultaneously maximize the number of leading ones and trailing zeroes.

\textsc{OneMinMax} has the property that every single solution represents a point in the Pareto front and that no search point is strictly dominated by another one. The goal is to cover the whole Pareto front, i.\,e.\ to compute a set of individuals that contains for each $i$, $0 \le i \le n$, an individual with exactly $i$ ones.
In the case of \textsc{LOTZ}, all non-Pareto optimal decision vectors only have Hamming neighbors that are better or worse, but never incomparable to it. This fact facilitates the analysis of the population-based algorithms, which certainly cannot be expected from other multi-objective optimization problems. Note that the Pareto front for \textsc{LOTZ} is given by the set of $n+1$ search points $\{1^i 0^{n-i} \mid 0 \le i \le n\}$.

We consider the Simple Evolutionary Multiobjective Optimizer (SEMO), shown in Algorithm~\ref{ds:alg:semo}, which is popular for theoretical analyses due to its simplicity. The work~\cite{CovantesOsuna2017a} also contains results for a variant GSEMO, which uses standard bit mutations instead of local mutations. For simplicity, we only present results for SEMO in this survey.

\begin{algorithm2e}[htb]
  	Choose an initial solution $s\in\{0,1\}^n$ uniformly at random.\\
    Determine $f(s)$ and initialize $P:=\{s\}$.
    \While{not stopping}{
    	Choose $s$ uniformly at random from $P$.\\
        Choose $i \in \{1,\ldots,n\}$ uniformly at random.\\
        Define $s'$ by flipping the $i$-th bit of $s$.\\
        \If{$s'$ is not dominated by any individual in $P$}{
        	Add $s'$ to $P$, and remove all individuals weakly dominated by $s'$ from~$P$.
        }
    }
  \caption{SEMO}
  \label{ds:alg:semo}
\end{algorithm2e}

The following theorem summarises results from~\cite{CovantesOsuna2017a,Giel2010,Laumanns2004a} on the performance of SEMO.
\begin{theorem}
\label{ds:the:semo_oneminmax}
The expected time for SEMO to cover the whole Pareto front on \textsc{OneMinMax} and \textsc{LOTZ} is $\Theta(n^2 \log n)$ and $\Theta(n^3)$, respectively.
%
\end{theorem}
The expected time is by a factor of $\Theta(n)$ larger than the expected time of the \dsea to optimize any single objective. The reason for SEMO being slower is that, once the Pareto front has been reached, only the search points with a maximum objective value, when chosen as parents, can expand the Pareto front further. All other choices of parents lead to the creation of an offspring whose objective values are already represented in the population.
Once the population has grown to a linear size~$\mu = \Theta(n)$, the probability of selecting a parent that allows SEMO to progress is only $\Theta(1/n)$, i.\,e.\ most steps are wasted. This leads to the additional factor of order~$n$ compared to the \dsea.

Diversity-based parent selection using either hypervolume contribution or crowding distance contribution can improve these running times. The parent selection mechanisms considered use one of these diversity metrics to select parents according to processes that favor higher diversity: sorting the population according to ranks of the diversity metric and picking the $i$-th ranked individual with probability proportional to $2^{-i}$ (exponential scheme) or $1/i^2$ (power-law scheme), or using tournament selection based on the diversity score with tournament size~$\mu$ (i.\,e.\ the current size of the population). Here the tournament is picked with replacement, hence search points can be picked multiple times, or be excluded from the tournament.
\begin{theorem}
\label{ds:the:semo_oneminmax-diversity}
Consider diversity-based parent selection using either hypervolume contribution or crowding distance contribution and selecting parents according to the exponential or power-law scheme, or according to a tournament with tournament size~$\mu$.

Then the expected time for SEMO with diversity-based parent selection to cover the whole Pareto front on \textsc{OneMinMax} and \textsc{LOTZ} is $O(n \log n)$ and $O(n^2)$, respectively.
\end{theorem}
The proofs show that the expected time for SEMO is bounded from above by $O((n \log n)/p_{\mathrm{good}})$ and $O(n^2/p_{\mathrm{good}})$, respectively, where $p_{\mathrm{good}}$ is (a lower bound on) the probability of selecting a parent that has a Hamming neighbor whose objective vector is on the Pareto front, but not yet represented in the population.
The diversity score assigns the highest values to the search points with maximum objective values.
However, the extreme points $0^n$ and $1^n$ may themselves not be ``good'' search points; if the Pareto front has reached one ``end'' of the search space, SEMO still may need to expand in the other direction. All mentioned parent selection mechanisms have a probability of $\Omega(1)$ for selecting the individual with the highest diversity rank, but they also have a probability of $\Omega(1)$ of selecting the second-best (and third-best) individual. Hence, even if the population does contain $0^n$ or $1^n$, the parent selection is still able to find a ``good'' parent to expand the Pareto front efficiently. Hence $p_{\mathrm{good}} = \Omega(1)$ and the claimed bounds follow.

\section{Conclusions}
\label{ds:sec:conclusions}

Maintaining and promoting diversity in evolutionary algorithms is a very important task. Surveys on diversity mechanisms~\cite{Shir2012,Squillero2016,Crepinsek2013} reveal a multitude of approaches to enhance and promote diversity, yet it is often unclear which of these mechanisms perform well, and why.

We have surveyed rigorous runtime analyses of evolutionary algorithms with explicit diversity mechanisms, ranging from avoiding genotype or fitness duplicates, deterministic crowding, fitness sharing and clearing to island models.
Other studies have shown that diversity can also emerge naturally, without any explicit mechanisms, through independent mutations, phases of independent evolution in the context of island models, or, in the case of \textsc{Jump}$_k$, through the interplay of different operators such as crossover followed by mutation and selection.

We have seen that diversity can be highly beneficial for enhancing the global exploration capabilities of evolutionary algorithms. It can enable crossover to work effectively, improve performance and robustness in dynamic optimization, and it is vital for evolutionary multiobjective optimization. In many cases diversity mechanisms can be highly effective for the considered problems, speeding up the expected or typical optimization time by constant factors, polynomial factors, or even exponential factors.

Comparing results for \twomax, \textsc{Jump}$_k$, and \textsc{Balance}, we found that diversity mechanisms that are effective for one problem may be ineffective for other problems, and vice versa. The analyses have rigorously quantified performance to demonstrate these effects. More importantly, they have laid the foundation for a rigorous understanding of how search dynamics are affected by the presence or absence of population diversity and introduction of diversity mechanisms.


\subsection*{Acknowledgments}

The author would like to thank Edgar Covantes Osuna for helpful comments. This work originated from Dagstuhl seminar 17191 ``Theory of Randomized Optimization Heuristics''; the author would like to thank the organisers and participants for inspiring discussions.

\bibliographystyle{abbrv}
\bibliography{diversity}

\end{document}

%% file: editors-macros.tex
\usepackage{url}

\usepackage[english]{babel}

\usepackage[utf8]{inputenc}
\usepackage{xspace}
\usepackage{amsmath,amssymb,mathtools}
\usepackage{lmodern}

\usepackage[algo2e,ruled,vlined,linesnumbered]{algorithm2e}

\usepackage{xcolor}
\usepackage{tikz}
\usepackage{graphicx}

\allowdisplaybreaks[4]
\newcommand{\om}{\textsc{OneMax}\xspace}
\newcommand{\onemax}{\om}

\newcommand{\R}{\ensuremath{\mathbb{R}}}

\newcommand{\N}{\ensuremath{\mathbb{N}}}



%% file: dirks-macros.tex
\usetikzlibrary{calc}
\usetikzlibrary{mindmap}
\usetikzlibrary{arrows}
\usetikzlibrary{shadows}
\usetikzlibrary{plotmarks}
\usetikzlibrary{backgrounds}
\usetikzlibrary{shapes}
\usetikzlibrary{shapes.symbols}
\usepackage{pgfplots}

\tikzstyle{helpline}=[black,thick]
\tikzstyle{curve}=[blue,thick]
\tikzstyle{grid}=[gray!50,very thin]
\tikzstyle{cut}=[helpline, blue]
\tikzstyle{vertex}=[fill=red,black,thick]
\tikzstyle{edge}=[black,semithick]
\tikzstyle{object}=[thick,draw=black,rounded corners]
\tikzstyle{knapsack}=[very thick,draw=brown,fill=brown!15]

\tikzstyle{algoarrow}=[very thick,-triangle 45,rounded corners]
\tikzstyle{algo operator}=[rectangle, rounded corners, draw, thick, inner sep=2mm, fill=green!60, drop shadow]
\tikzstyle{algo operator alerted}=[algo operator, fill=red!60]
\tikzstyle{algo decision}=[diamond, draw, thick, inner sep=1.5mm, fill=gray!60, drop shadow]

\tikzstyle{uninformed}=[green!60]
\tikzstyle{informed1}=[red!80]
\tikzstyle{informed2}=[red!30]
\tikzstyle{informed3}=[red!15]

\newcommand{\twomax}{\textsc{TwoMax}\xspace}
\newcommand{\Twomax}{\twomax}

\newcommand{\dsea}{(1+1)~EA\xspace}
\newcommand{\dsmuea}{($\mu$+1)~EA\xspace}
\newcommand{\dstworls}{(2+1)~RLS\xspace}
\newcommand{\dsmlea}{($\mu$+$\lambda$)~EA\xspace}
\newcommand{\dsmuGA}{($\mu$+$1$)~GA\xspace}
\newcommand{\dsmlGA}{($\mu$+$\lambda$)~GA\xspace}
\newcommand{\numislands}{m}
\newcommand{\numproc}{\lambda}
\newcommand{\migint}{\tau}

\DeclareMathOperator{\diam}{diam}

%% file: diversity-arxiv.bbl
\begin{thebibliography}{10}

\bibitem{Barton:2013:QPG:2463372.2463568}
N.~Barton and T.~Paix\~{a}o.
\newblock Can quantitative and population genetics help us understand
  evolutionary computation?
\newblock In {\em Proceedings of the Genetic and Evolutionary Computation
  Conference (GECCO~'13)}, pages 1573--1580, 2013.

\bibitem{Bleuler2001}
S.~Bleuler, M.~Brack, L.~Thiele, and E.~Zitzler.
\newblock Multiobjective genetic programming: reducing bloat using {SPEA2}.
\newblock In {\em Proceedings of the 2001 Congress on Evolutionary Computation
  (CEC 2001)}, volume~1, pages 536--543, 2001.

\bibitem{Brockhoff2011}
D.~Brockhoff.
\newblock Theoretical aspects of evolutionary multiobjective optimization.
\newblock In {\em Theory of Randomized Search Heuristics--Foundations and
  Recent Developments}. World Scientific Publishing, 2011.

\bibitem{Corus2017b}
D.~Corus and P.~S. Oliveto.
\newblock Standard steady state genetic algorithms can hillclimb faster than
  mutation-only evolutionary algorithms.
\newblock {\em IEEE Transactions on Evolutionary Computation}, 2017.
\newblock To appear.

\bibitem{CovantesOsuna2017a}
E.~Covantes~Osuna, W.~Gao, F.~Neumann, and D.~Sudholt.
\newblock Speeding up evolutionary multi-objective optimisation through
  diversity-based parent selection.
\newblock In {\em Proceedings of the Genetic and Evolutionary Computation
  Conference (GECCO '17)}, pages 553--560. ACM, 2017.

\bibitem{CovantesOsuna2017}
E.~Covantes~Osuna and D.~Sudholt.
\newblock Analysis of the clearing diversity-preserving mechanism.
\newblock In {\em Proceedings of Foundations of Genetic Algorithms (FOGA
  2017)}, pages 55--63. ACM Press, 2017.

\bibitem{Dang2017}
D.-C. Dang, T.~Friedrich, T.~K{\"o}tzing, M.~S. Krejca, P.~K. Lehre, P.~S.
  Oliveto, D.~Sudholt, and A.~M. Sutton.
\newblock Escaping local optima using crossover with emergent diversity.
\newblock {\em IEEE Transactions on Evolutionary Computation}.
\newblock To appear.

\bibitem{Dang2016}
D.-C. Dang, T.~Friedrich, M.~S. Krejca, T.~K{\"o}tzing, P.~K. Lehre, P.~S.
  Oliveto, D.~Sudholt, and A.~M. Sutton.
\newblock Escaping {{Local Optima}} with {{Diversity}}-{{Mechanisms}} and
  {{Crossover}}.
\newblock In {\em Proceedings of the Genetic and Evolutionary Computation
  Conference ({GECCO} 2016)}, pages 645--652. {ACM Press}.

\bibitem{Dang2017a}
D.-C. Dang, T.~Jansen, and P.~K. Lehre.
\newblock Populations can be essential in tracking dynamic optima.
\newblock {\em Algorithmica}, 78(2):660--680, Jun 2017.

\bibitem{DeFelice2011}
M.~De~Felice, S.~Meloni, and S.~Panzieri.
\newblock Effect of topology on diversity of spatially-structured evolutionary
  algorithms.
\newblock In {\em Proceedings of the Genetic and Evolutionary Computation
  Conference (GECCO~'11)}, pages 1579--1586. ACM, 2011.

\bibitem{Deb2002}
K.~Deb, A.~Pratap, S.~Agarwal, and T.~Meyarivan.
\newblock A fast and elitist multiobjective genetic algorithm: {NSGA-II}.
\newblock {\em IEEE Transactions on Evolutionary Computation}, 6(2):182--197,
  Apr 2002.

\bibitem{Doerr2012}
B.~Doerr, E.~Happ, and C.~Klein.
\newblock Crossover can provably be useful in evolutionary computation.
\newblock {\em Theoretical Computer Science}, 425(0):17--33, 2012.

\bibitem{Doerr2007}
B.~Doerr, N.~Hebbinghaus, and F.~Neumann.
\newblock Speeding up evolutionary algorithms through asymmetric mutation
  operators.
\newblock {\em Evolutionary Computation}, 15:401--410, 2007.

\bibitem{Doerr2007a}
B.~Doerr and D.~Johannsen.
\newblock Adjacency list matchings---an ideal genotype for cycle covers.
\newblock In {\em Proceedings of the Genetic and Evolutionary Computation
  Conference (GECCO~'07)}, pages 1203--1210. ACM Press, 2007.

\bibitem{Doerr2007e}
B.~Doerr, C.~Klein, and T.~Storch.
\newblock Faster evolutionary algorithms by superior graph representation.
\newblock In {\em First IEEE Symposium on Foundations of Computational
  Intelligence (FOCI~'07)}, pages 245--250. IEEE, 2007.

\bibitem{Fischer2005}
S.~Fischer and I.~Wegener.
\newblock The one-dimensional {I}sing model: {M}utation versus recombination.
\newblock {\em Theoretical Computer Science}, 344(2--3):208--225, 2005.

\bibitem{Friedrich2007}
T.~Friedrich, N.~Hebbinghaus, and F.~Neumann.
\newblock Rigorous analyses of simple diversity mechanisms.
\newblock In {\em Proceedings of the Genetic and Evolutionary Computation
  Conference (GECCO~'07)}, pages 1219--1225. ACM Press, 2007.

\bibitem{Friedrich2009}
T.~Friedrich, P.~S. Oliveto, D.~Sudholt, and C.~Witt.
\newblock Analysis of diversity-preserving mechanisms for global exploration.
\newblock {\em Evolutionary Computation}, 17(4):455--476, 2009.

\bibitem{Gao2014}
W.~Gao and F.~Neumann.
\newblock Runtime analysis for maximizing population diversity in
  single-objective optimization.
\newblock In {\em Proc.\ of GECCO~'14}, pages 777--784, 2014.

\bibitem{Giacobini2005}
M.~Giacobini, M.~Tomassini, and A.~Tettamanzi.
\newblock Takeover time curves in random and small-world structured
  populations.
\newblock In {\em Proceedings of the Genetic and Evolutionary Computation
  Conference (GECCO~'05)}, pages 1333--1340. ACM Press, 2005.

\bibitem{Giel2010}
O.~Giel and P.~K. Lehre.
\newblock On the effect of populations in evolutionary multi-objective
  optimisation.
\newblock {\em Evolutionary Computation}, 18(3):335--356, Sept. 2010.

\bibitem{HorobaJansenZarges09}
C.~Horoba, T.~Jansen, and C.~Zarges.
\newblock Maximal age in randomized search heuristics with aging.
\newblock In {\em Proceedings of the Genetic and Evolutionary Computation
  Conference (GECCO~'09)}, pages 803--810, 2009.

\bibitem{Horoba2010b}
C.~Horoba and F.~Neumann.
\newblock {\em Approximating Pareto-Optimal Sets Using Diversity Strategies in
  Evolutionary Multi-Objective Optimization}, pages 23--44.
\newblock Springer Berlin Heidelberg, 2010.

\bibitem{Hutter2006}
M.~Hutter and S.~Legg.
\newblock Fitness uniform optimization.
\newblock {\em IEEE Transactions on Evolutionary Computation}, 10:568--589,
  2006.

\bibitem{Jansen2002}
T.~Jansen and I.~Wegener.
\newblock On the analysis of evolutionary algorithms---a proof that crossover
  really can help.
\newblock {\em Algorithmica}, 34(1):47--66, 2002.

\bibitem{Jansen2005c}
T.~Jansen and I.~Wegener.
\newblock Real royal road functions---where crossover provably is essential.
\newblock {\em Discrete Applied Mathematics}, 149:111--125, 2005.

\bibitem{JansenZarges2011a}
T.~Jansen and C.~Zarges.
\newblock Analyzing different variants of immune inspired somatic contiguous
  hypermutations.
\newblock {\em Theoretical Computer Science}, 412(6):517--533, 2011.

\bibitem{JansenZarges2011c}
T.~Jansen and C.~Zarges.
\newblock On the role of age diversity for effective aging operators.
\newblock {\em Evolutionary Intelligence}, 4(2):99--125, 2011.

\bibitem{Jansen2014a}
T.~Jansen and C.~Zarges.
\newblock Evolutionary algorithms and artificial immune systems on a bi-stable
  dynamic optimisation problem.
\newblock In {\em Proceedings of the 2014 Annual Conference on Genetic and
  Evolutionary Computation (GECCO~'14)}, pages 975--982. ACM, 2014.

\bibitem{Komarova}
N.~L. Komarova, E.~Urwin, and D.~Wodarz.
\newblock Accelerated crossing of fitness valleys through division of labor and
  cheating in asexual populations.
\newblock {\em Scientific Reports}, 2012.

\bibitem{Kotzing2015}
T.~K\"{o}tzing, A.~Lissovoi, and C.~Witt.
\newblock (1+1) {EA} on generalized dynamic onemax.
\newblock In {\em Proceedings of the 2015 ACM Conference on Foundations of
  Genetic Algorithms (FOGA~'15)}, pages 40--51. ACM, 2015.

\bibitem{Koetzing2012b}
T.~K{\"o}tzing and H.~Molter.
\newblock {ACO} beats {EA} on a dynamic pseudo-boolean function.
\newblock In {\em Parallel Problem Solving from Nature (PPSN XII)}, pages
  113--122. Springer Berlin Heidelberg, 2012.

\bibitem{Koetzing2011a}
T.~K{\"o}tzing, D.~Sudholt, and M.~Theile.
\newblock How crossover helps in pseudo-{B}oolean optimization.
\newblock In {\em Proceedings of the 13th Annual Genetic and Evolutionary
  Computation Conference (GECCO~2011)}, pages 989--996. ACM Press, 2011.

\bibitem{Lassig2010}
J.~L{\"a}ssig and D.~Sudholt.
\newblock The benefit of migration in parallel evolutionary algorithms.
\newblock In {\em Proceedings of the Genetic and Evolutionary Computation
  Conference (GECCO 2010)}, pages 1105--1112. ACM Press, 2010.

\bibitem{Lassig2013}
J.~L{\"a}ssig and D.~Sudholt.
\newblock Design and analysis of migration in parallel evolutionary algorithms.
\newblock {\em Soft Computing}, 17(7):1121--1144, 2013.

\bibitem{Lassig2014}
J.~L{\"a}ssig and D.~Sudholt.
\newblock Analysis of speedups in parallel evolutionary algorithms and
  (1+$\lambda$)~{EAs} for combinatorial optimization.
\newblock {\em Theoretical Computer Science}, 551:66--83, 2014.

\bibitem{Laumanns2004a}
M.~Laumanns, L.~Thiele, and E.~Zitzler.
\newblock Running time analysis of multiobjective evolutionary algorithms on
  pseudo-boolean functions.
\newblock {\em IEEE Transactions on Evolutionary Computation}, 8(2):170--182,
  April 2004.

\bibitem{Lissovoi2015}
A.~Lissovoi and C.~Witt.
\newblock On the utility of island models in dynamic optimization.
\newblock In {\em Proceedings of the 2015 Annual Conference on Genetic and
  Evolutionary Computation}, GECCO '15, pages 1447--1454, New York, NY, USA,
  2015. ACM.

\bibitem{Lissovoi2017a}
A.~Lissovoi and C.~Witt.
\newblock A runtime analysis of parallel evolutionary algorithms in dynamic
  optimization.
\newblock {\em Algorithmica}, 78(2):641--659, 2017.

\bibitem{Mahfoud1997}
S.~W. Mahfoud.
\newblock Niching methods.
\newblock In T.~B{\"a}ck, D.~B. Fogel, and Z.~Michalewicz, editors, {\em
  Handbook of {E}volutionary {C}omputation}, pages C6.1:1--4. Institute of
  Physics Publishing and Oxford University Press, Bristol, New York, 1997.

\bibitem{Mambrini2012}
A.~Mambrini, D.~Sudholt, and X.~Yao.
\newblock Homogeneous and heterogeneous island models for the set cover
  problem.
\newblock In {\em Parallel Problem Solving from Nature (PPSN 2012)}, volume
  7491 of {\em LNCS}, pages 11--20. Springer, 2012.

\bibitem{Neumann2008d}
F.~Neumann.
\newblock {Expected runtimes of evolutionary algorithms for the Eulerian cycle
  problem}.
\newblock {\em Computers \& Operations Research}, 35(9):2750--2759, 2008.

\bibitem{Neumann2011}
F.~Neumann, P.~S. Oliveto, G.~Rudolph, and D.~Sudholt.
\newblock On the effectiveness of crossover for migration in parallel
  evolutionary algorithms.
\newblock In {\em Proceedings of the Genetic and Evolutionary Computation
  Conference (GECCO~2011)}, pages 1587--1594. ACM Press, 2011.

\bibitem{Oliveto2008tr}
P.~S. Oliveto, J.~He, and X.~Yao.
\newblock Population-based evolutionary algorithms for the vertex cover
  problem.
\newblock In {\em Proceedings of the IEEE Congress on Evolutionary Computation
  (CEC~'08)}, pages 1563--1570, 2008.

\bibitem{Oliveto2014}
P.~S. Oliveto and D.~Sudholt.
\newblock On the runtime analysis of stochastic ageing mechanisms.
\newblock In {\em Proceedings of the Genetic and Evolutionary Computation
  Conference (GECCO 2014)}, pages 113--120. ACM Press, 2014.

\bibitem{Oliveto2014a}
P.~S. Oliveto, D.~Sudholt, and C.~Zarges.
\newblock On the runtime analysis of fitness sharing mechanisms.
\newblock In {\em 13th International Conference on Parallel Problem Solving
  from Nature (PPSN 2014)}, volume 8672 of {\em LNCS}, pages 932--941.
  Springer, 2014.

\bibitem{Oliveto2015}
P.~S. Oliveto and C.~Zarges.
\newblock Analysis of diversity mechanisms for optimisation in dynamic
  environments with low frequencies of change.
\newblock {\em Theoretical Computer Science}, 561:37--56, 2015.

\bibitem{RohlfshagenLehreYao2009}
P.~Rohlfshagen, P.~K. Lehre, and X.~Yao.
\newblock Dynamic evolutionary optimisation: an analysis of frequency and
  magnitude of change.
\newblock In {\em Proceedings of the 2009 Genetic and Evolutionary Computation
  Conference (GECCO '09)}, pages 1713--1720. ACM Press, 2009.

\bibitem{Shir2012}
O.~M. Shir.
\newblock Niching in evolutionary algorithms.
\newblock In G.~Rozenberg, T.~B{\"a}ck, and J.~N. Kok, editors, {\em Handbook
  of Natural Computing}, pages 1035--1070. Springer, 2012.

\bibitem{Squillero2016}
G.~Squillero and A.~Tonda.
\newblock Divergence of character and premature convergence: A survey of
  methodologies for promoting diversity in evolutionary optimization.
\newblock {\em Information Sciences}, 329:782--799, 2016.
\newblock Special issue on Discovery Science.

\bibitem{Storch2004}
T.~Storch and I.~Wegener.
\newblock Real royal road functions for constant population size.
\newblock {\em Theoretical Computer Science}, 320:123--134, 2004.

\bibitem{Sudholt2016}
D.~Sudholt.
\newblock How crossover speeds up building-block assembly in genetic
  algorithms.
\newblock {\em Evolutionary Computation}, 25(2):237--274.

\bibitem{Sudholt2005}
D.~Sudholt.
\newblock Crossover is provably essential for the {Ising} model on trees.
\newblock In {\em Proceedings of the Genetic and Evolutionary Computation
  Conference (GECCO~'05)}, pages 1161--1167. ACM Press, 2005.

\bibitem{Sudholt2010}
D.~Sudholt.
\newblock Hybridizing evolutionary algorithms with variable-depth search to
  overcome local optima.
\newblock {\em Algorithmica}, 59(3):343--368, 2011.

\bibitem{Sudholt2012b}
D.~Sudholt.
\newblock Crossover speeds up building-block assembly.
\newblock In {\em Proceedings of the Genetic and Evolutionary Computation
  Conference (GECCO 2012)}, pages 689--696. ACM Press, 2012.

\bibitem{Sudholt2012a}
D.~Sudholt.
\newblock Parallel evolutionary algorithms.
\newblock In J.~Kacprzyk and W.~Pedrycz, editors, {\em Handbook of
  Computational Intelligence}, pages 929--959. Springer, 2015.

\bibitem{Crepinsek2013}
M.~\v{C}repin\v{s}ek, S.-H. Liu, and M.~Mernik.
\newblock Exploration and exploitation in evolutionary algorithms: A survey.
\newblock {\em ACM Computing Surveys}, 45(3):35:1--35:33, 2013.

\bibitem{Watson2007}
R.~A. Watson and T.~Jansen.
\newblock A building-block royal road where crossover is provably essential.
\newblock In {\em Proceedings of the Genetic and Evolutionary Computation
  Conference (GECCO~'07)}, pages 1452--1459. ACM, 2007.

\bibitem{Wegener2002}
I.~Wegener.
\newblock Methods for the analysis of evolutionary algorithms on
  pseudo-{Boolean} functions.
\newblock In R.~Sarker, X.~Yao, and M.~Mohammadian, editors, {\em Evolutionary
  Optimization}, pages 349\protect\nobreakdash--369. Kluwer, 2002.

\bibitem{Weissman1389}
D.~B. Weissman, M.~W. Feldman, and D.~S. Fisher.
\newblock The rate of fitness-valley crossing in sexual populations.
\newblock {\em Genetics}, 186:1389--1410, 2010.

\bibitem{Witt2006}
C.~Witt.
\newblock Runtime analysis of the {($\mu$+1) EA} on simple pseudo-{Boolean}
  functions.
\newblock {\em Evolutionary Computation}, 14(1):65--86, 2006.

\bibitem{Zarges2017}
C.~Zarges.
\newblock {\em Theoretical Foundations of Immune-Inspired Randomized Search
  Heuristics for Optimisation}.

\bibitem{Zitzler2004}
E.~Zitzler and S.~K{\"u}nzli.
\newblock Indicator-based selection in multiobjective search.
\newblock In {\em Proceedings of the Parallel Problem Solving from Nature -
  PPSN VIII}, pages 832--842. Springer Berlin Heidelberg, 2004.

\end{thebibliography}
